\documentclass[journal,twoside,web]{ieeecolor}

\usepackage{etoolbox}
\makeatletter
\@ifundefined{color@begingroup}%
{\let\color@begingroup\relax
\let\color@endgroup\relax}{}%
\def\fix@ieeecolor@hbox#1{%
\hbox{\color@begingroup#1\color@endgroup}}
\patchcmd\@makecaption{\hbox}{\fix@ieeecolor@hbox}{}{\FAILED}
\patchcmd\@makecaption{\hbox}{\fix@ieeecolor@hbox}{}{\FAILED}
\usepackage{multirow}
\usepackage{tmi}
\usepackage{cite}
\usepackage{amsmath,amssymb,amsfonts}
\usepackage{algorithmic}
\usepackage{graphicx}
\usepackage{textcomp}
\def\BibTeX{{\rm B\kern-.05em{\sc i\kern-.025em b}\kern-.08em
    T\kern-.1667em\lower.7ex\hbox{E}\kern-.125emX}}
\begin{document}
\title{Attention-based Shape-Deformation Networks for Artifact-Free Geometry Reconstruction of Lumbar Spine from MR Images}
\author{Linchen Qian, Jiasong Chen, Linhai Ma, Timur Urakov, Weiyong Gu, Liang Liang
%First A. Author, \IEEEmembership{Fellow, IEEE}, Second B. Author,and Third C. Author, Jr., \IEEEmembership{Member, IEEE}
%\thanks{This paragraph of the first footnote will contain the date on which
%you submitted your paper for review. It will also contain support information,
%including sponsor and financial support acknowledgment. For example, 
%``This work was supported in part by the U.S. Department of Commerce under Grant BS123456.'' }
\thanks{Linchen Qian, Jiasong Chen, Linhai Ma, and Liang Liang are with
the Department of Computer Science, College of Arts and Sciences, University of Miami, Coral Gables, FL 33146, USA
(e-mail: linchen.qian@miami.edu; jiasong.chen@miami.edu; l.ma@miami.edu; liang@cs.miami.edu).}
\thanks{Timur Urakov is with the Department of Neurological Surgery, University of Miami Miller School of Medicine, Miami, FL 33136, USA (e-mail: turakov@med.miami.edu).}
\thanks{Weiyong Gu is with the Department of Mechanical and Aerospace Engineering, College of Engineering, University of Miami, Coral Gables, FL 33146, USA (e-mail: wgu@miami.edu).}}

\maketitle

\begin{abstract}
Lumbar disc degeneration, a progressive structural wear and tear of lumbar intervertebral disc, is regarded as an essential role on low back pain, a significant global health concern. % Medical parameters related to the lumbar spine geometry features to evaluate the lumbar status, 
Automated lumbar spine geometry reconstruction from MR images will enable fast measurement of medical parameters to evaluate the lumbar status, in order to determine a suitable treatment. % can model the topology and perform quantitative analysis of the degeneration cascade
Existing image segmentation-based techniques often generate erroneous segments or unstructured point clouds, unsuitable for medical parameter measurement. % without point correspondence between patients.
In this work, we present \textit{UNet-DeformSA} and \textit{TransDeformer}: novel attention-based deep neural networks that reconstruct the geometry of the lumbar spine with high spatial accuracy and mesh correspondence across patients, and we also present a variant of \textit{TransDeformer} for error estimation.
Specially, we devise new attention modules with a new attention formula, which integrate image features and tokenized contour features to predict the displacements of the points on a shape template without the need for image segmentation.
The deformed template reveals the lumbar spine geometry in an image.
%We develop a multi-stage training strategy to enhance model robustness with respect to template initialization.
Experiment results show that our networks generate artifact-free geometry outputs, and the variant of \textit{TransDeformer} can predict the errors of a reconstructed geometry.
% Also, we propose a multi-stage training strategy to enable stages to specialize in different aspects of point-specific displacements, thereby enhancing the accuracy and robustness of geometry reconstruction.
% The contours remain point correspondence between patients providing intuitive medical measures for assessing key aspects of the spinal components without any manual post-processing.
Our code is available at https://github.com/linchenq/TransDeformer-Mesh.

\end{abstract}

\begin{IEEEkeywords}
% Enter about five key words or phrases in alphabetical order, separated by commas.
deep learning, geometry reconstruction, lumbar spine, mesh correspondence, attention
\end{IEEEkeywords}

\section{Introduction}
\label{sec:introduction}
\IEEEPARstart{G}{eometry} reconstruction of the anatomical structures from medical images helps to improve clinical outcomes such as disease diagnosis accuracy, surgical planning accuracy, and treatment efficacy \cite {b46, b47}.
Low back pain is a prevalent global health issue associated with activity limitation and absenteeism from work \cite{b7, b9}, and lumbar disc degeneration, which is the gradual deterioration of the lumbar intervertebral disc, plays a significant role in the onset of this issue \cite{b8}.
Magnetic Resonance Imaging (MRI) is instrumental in identifying morphological changes and revealing the internal structure of tissues, which is recognized as a key method for investigating disc degeneration \cite{b10}.
Consequently, it is essential to model lumbar spine morphology and conduct a quantitative analysis of degeneration cascade using a precise geometric representation \cite{b11} that is reconstructed from lumbar spine MR images.
Due to individual variations, manual geometry annotation is laborious, and therefore automated image analysis methods are desired for clinical applications.

\begin{figure}[htbp]
% Syntax: \includegraphics[trim={left bottom right top},clip]{filename}
\centerline{\includegraphics[width=\columnwidth]{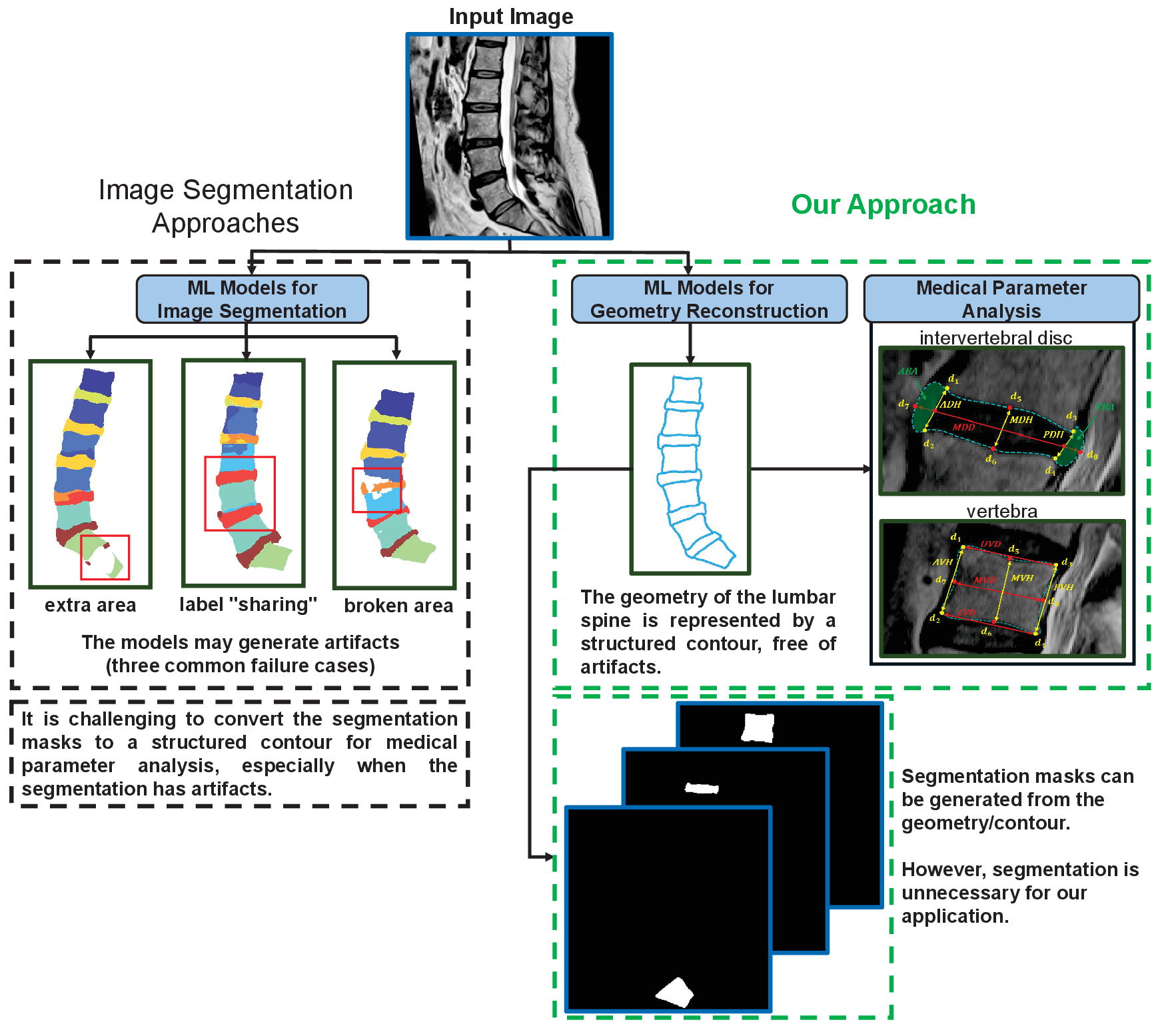}}
\caption{Comparison of our geometry reconstruction approach (right)  and the existing image segmentation approaches (left) for lumbar spine MR image analysis.}
\label{fig:intro-geometry}
\end{figure}

Currently, most of the lumbar spine image analysis applications focus on image segmentation and classification \cite{b45}.
%\cite{b4, b5, b3, b45}. 
The object masks from image segmentation are not directly usable for measuring many medical parameters related to lumbar disc degeneration in our application, and those masks have to be converted to a mesh, on which the medical parameters could be defined and measured. The mask-to-mesh conversion can be error-prone because segmentation artifacts are often present in the segmentation masks, as illustrated in Fig. \ref{fig:intro-geometry}.

%Whereas numerous studies focus on image segmentation, the informative geometric representation (e.g., a mesh or contour), particularly in the context of lumbar spine, offers intuitive measurements and consistent point correspondence across patients \cite{b4, b5}.
In contrast to segmentation masks, a mesh (i.e., a structured contour) of the lumbar spine is a more precise representation that depicts the anatomy with nodes/points and elements (i.e., point connectives), as shown in  Fig. \ref{fig:intro-geometry}.
%only coordinates of a contour's points stored \cite{b1}.
This mesh representation facilitates the computation of the medical parameters that reflect disc degeneration status \cite{b11, b13}.
Moreover, if meshes with the same topology are used for all patients, mesh correspondence across patients will be established, thus supporting consistent geometric analysis using the same reference anatomical structure \cite{b12}. 
Despite the advantages over segmentation masks, automated geometry reconstruction of the lumbar spine with a mesh representation remains a challenge because of anatomical complexity and variations.

In this study, we propose two new deep neural networks, named \textbf{\textit{UNet-DeformSA}} and \textbf{\textit{TransDeformer}}, to reconstruct lumbar spine geometries from 2D sagittal MR images by deforming a mesh template, thereby enabling consistent definition and measurement of the medical parameters related to lumbar disc degeneration. By using new attention modules with a new attention formula, the geometry outputs of our networks are artifact-free. In addition, we propose a third network to estimate the errors of the geometries reconstructed by a network, which facilitates quality control for clinical use.

Due to the 10-page limit, we keep the following sections to be as concise as possible. We can expand the reference list if the reviewers consider it necessary.

%We designed two new deep neural networks by utilizing the attention mechanisms \cite{b19}. Our networks deform a pre-defined template according to the image features in the MR image of a patient, and the deformed template is the reconstructed geometry, which ensures a mesh representation with a pre-defined topology/structure. This is the first study on exploring the potential of attention mechanisms for geometry reconstruction of lumbar spine from MR images using a template. 

%from the predefined template without any manual post-processing.
%$Our model, dubbed TransDeformer (\textbf{Tran}sformer-based \textbf{S}hape \textbf{Deform}ation), leverages the transformer that integrates extracted features from MRI images and tokenized contour features to predict precise displacements of the points on the shape template.
%Also, the model utilizes image features of varying resolutions to enable refinement of contours on multiple scales, and eliminates additional supervision by segmentation unlike previous approaches.
%It demonstrates superior spatial accuracy and avoids abnormal casses (e.g. extraneous, overlapping, or missing regions).

\section{Related Works}
\label{sec:related-works}
%rewrite this section
%(1) very brief review of the image segmentation models, and state that segmentation masks from these models may have artifacts
%(2) template-deforming based geometry Reconstruction methods

%The geometry of an object is represented by different formats in different tasks, such as a segmentation mask, a point cloud, or a mesh \cite{b20}.
%In our study, we define the geometry of the lumbar spine as the interconnected contours of $6$ vertebrae and $5$ intervertebral discs, which is an organized set of point coordinates that have consistent correspondence where the same anatomical structure is maintained across different patients.
%This simplifies further quantitative analysis by providing a consistent and reliable representation of the lumbar spine.
%Also, the geometry, with point coordinates systematically stored, enables straightforward transformation to segmentation (e.g., surrounding areas) but not vice versa (Fig. \ref{fig:intro-geometry}). 

In our application, the geometry of a lumbar spine is represented by a 2D mesh (see Fig. \ref{fig:intro-geometry}), i.e., a structured contour of the 11 lumbar spine objects including 6 vertebrae and 5 intervertebral discs. Throughout this paper, lumbar spine geometry, shape, and contour are used interchangeably. 
%To enable consistent definition and measurement of the medical parameters as well as direct comparison between different shapes, the contour needs to have a fixed number of points and fixed connectivity between points. 

As briefly mentioned in the introduction section, in ideal scenarios, the object masks from an image segmentation model could serve as an intermediate representation, from which a mesh could be extracted. In a recent study \cite{b33}, we evaluated fifteen image segmentation models for our application, (e.g., UNet++ \cite{b40}, TransUnet \cite{b28}, Swin-Unet \cite{b41}, BianqueNet\cite{b5}). Although  some of these image segmentation models perform well with Dice scores $>$ 0.9 on average, segmentation artifacts are often present in the segmentation masks, which may be (1) extra areas not belonging to any discs or vertebrae, (2) the same class label assigned to two different discs, and (3) brokens area of a disc or vertebrae. A mesh extracted from the segmentation masks with artifacts has the same artifacts and therefore is useless. Thus, the segmentation-and-meshing approach do not work for our application.

A template deformation based approach for geometry reconstruction is a better choice for our application, because it ensures mesh correspondence across patients and thereby enables consistent definitions of medical parameters on a reference anatomical structure (i.e., the template). This type of approaches \cite{b26} can be traced back to a method known as active appearance model (AAM) \cite{b44} that deforms a shape template with the guidance of the statistical model of object geometry and appearance. It is known that the AAM method will not work if images have complex textures and weak edges. To handle complex data, in the pre-deep learning era, classic machine learning (ML) methods were used for template initialization and nonrigid deformation. For example, in the application of heart geometry modeling from CT images \cite{b42}, a template was initialized by an object detector and then deformed by using an object boundary detector \cite{b42}, where the detectors were built on probabilistic boosting-tree techniques. In the application of aortic valve leaflet geometry modeling from CT images \cite{b43}, a template was deformed by using linear coding with a dictionary of representative shapes.  
%[ref-A Four-chamber heart modeling and automatic segmentation for 3-D cardiac CT volumes using marginal space learning and steerable features]
%[ref-B Machine learning–based 3‐D geometry reconstruction and modeling of aortic valve deformation using 3‐D computed tomography images]
Deep learning (DL) has significantly changed the landscape of medical image analysis \cite{b47}.
%DL techniques, including convolutional neural networks (CNNs), graph convolutional networks (GCNs), and vision Transformers (ViTs), have demonstrated remarkable success in various tasks of medical image analysis. 
In the next few paragraphs, we provide a brief review of the existing medical DL approaches for geometry reconstruction with template deformation.

Graph convolution network (GCN) has demonstrated ground-breaking performance in geometry modeling by leveraging the structural relationships among vertices and edges \cite{b48}. 
%\cite{b16}. 
%The GCN transforms the geometry into graph structure where each vertex is assigned with learned features and each edge records the connections between vertices.
GCNs can predict displacements of points/vertices of a template using image features.
Wickramasinghe et al. proposed Voxel2Mesh, a GCN-based model with adaptive sampling and pooling strategies for geometry reconstruction of liver, hippocampus, and synaptic junction \cite{b2}.
Although this Voxel2Mesh approach outperformed several segmentation-based methods, it generated an uncertain number of vertices, thus lacking mesh correspondence across patients.
%, which requires post-processing for quantitative evaluation.
Kong et al. proposed a compound framework fusing a UNet with GCN to deform templates for whole-heart geometry reconstruction on cardiac datasets \cite{b1}.
%, which undertook image feature encoding and mesh template deformation to reconstruct the whole-heart geometry on public cardiac datasets \cite{b1}. 
As will be shown in our experiments, such a UNet-GCN approach still generates geometry artifacts in our application.
%The approach introduced ground truth segmentation as supervision to regularize the encoded image features, which were used by progressive GCN blocks to estimate the deviation between the shape template and the desired geometry. 
%Yet, this supervision by binary segmentation did not avoid fragmented artifacts, injecting unexpected disruption into the fused information of image features that led to abnormal geometry.

Instead of treating a geometry as a graph, some studies regard geometry reconstruction as a template-image registration task, wherein a diffeomorphic displacement field \cite{b21} is generated by a network to transform an initial template mesh to the target. 
For example, Pak et al. proposed DeepCarve that predicted a regular-grid displacement field by a UNet and used it to deform a pre-defined template of left-ventricle and aorta complex  on a dataset of 80 cardiac CT scans \cite{b20}.
%The approach also incorporated a mix of isotropic and anisotropic energies to preserve the quality of volumetric mesh. 
As will be shown in our experiments, such a UNet-Disp approach is very sensitive to template initialization, such that some perturbation on initial position of the template may aggravate the quality of the output geometry.
%Nevertheless, the acquired spatial transformation highly relied on the initial position of the template, such that minor perturbation on the template may aggravate the quality of the target geometry.

Recently, Transformers utilizing an attention mechanism, have been adopted for image classification and segmentation \cite{b49}.
%\cite{b24, b28}.
%What is the network used in Fig.1 for image segmentation? TransUNet? TransFusion?
The output of a scaled dot-product attention in Transformers is mathematically formulated as:
\begin{equation}
    Out\left(Q,K,V\right)=softmax\left(\frac{QK^{T}}{\sqrt{d_{k}}}\right)V
\label{eq:saeq}
\end{equation}
%Let $X_F$ be the input vector. Here $X_{F}\in{\mathbb{R}^{C_{in}\times{H}\times{W}}}$, where $\mathbb{R}$ is real number, $C_{in}$ is the channels of the input feature, $H$ and $W$ are spatial height and width.
Linear projections are applied to the input vectors, resulting in query ($Q$), key ($K$), and value ($V$) matrices. 
%Here, $Q$, $K$, and $V$ $\in{\mathbb{R}^{C_{out}\times{E}}}$ where $E$ is the embedding dimension. 
$d_k$ is the dimension of a query or key.
Unlike convolution which is akin to uniform filtering across different locations on the input, the attention mechanism in Transformers operates as an adaptive filter, where its weights are determined by interrelations between each paired tokens.
%Therefore, the attention mechanism provides a context-aware analysis across entire data sequence by using dynamically allocated weights on the input vectors.

In this paper, we present the first study on exploring the potential of attention mechanisms for geometry reconstruction of lumbar spine from MR images using a template. 

\section{Methods}
\label{sec:methods}

As illustrated by Fig. \ref{fig:intro-geometry}, the goal of our application is to reconstruct lumbar spine geometries from 2D sagittal MR images and then measure the medical parameters that are often used for lumbar disc degeneration assessment. For this purpose, the geometry of a lumbar spine is represented by a 2D mesh, i.e., a structured contour of the 11 lumbar spine components including 6 vertebrae and 5 intervertebral discs. 
%Throughout this paper, lumbar spine geometry, shape, and contour are used interchangeably.
%To enable consistent definition and measurement of the medical parameters as well as direct comparison between different shapes, the contour needs to have a fixed number of points and fixed connectivity between points. Therefore, a template deformation based approach for geometry reconstruction is the best choice for the application.

We designed two novel networks for lumbar spine geometry reconstruction, named \textbf{\textit{UNet-DeformSA}} and \textbf{\textit{TransDeformer}}. The UNet-DeformSA model (Fig. \ref{fig:unet-deformsa-model}) has a UNet backbone and a shape self-attention (SSA) mechanism that sets it apart from the existing models. The TransDeformer model (Fig. \ref{fig:net6}) gets rid of the UNet backbone by using cross-attention between shape and image as well as shape self-attention and image self-attention, which further improves the performance. The attention mechanisms in ViT and its successors \cite{b17} assume that tokens are located on a regular grid and therefore could not be used in our networks. To enable the new attention modules in our networks, we developed new equations for attention with relative position embedding. Furthermore, we modified TransDeformer for error estimation, i.e., to estimate the error of a reconstructed geometry. In the following sections, we provide the details of our models. For information not revealed due to the 10-page limit, we refer the reader to the source code.

\subsection{Attention with Relative Position Embedding}
To enable the new attention modules in our networks, we developed a new set of equations for attention with relative position embedding. Let $X$ and $Y$ represent two matrices, and each row of a matrix is a token. A token has a unique spatial position. The tokens in $X$ and $Y$ will have different meanings in different attention modules in our networks, which will be explained in other sections. $X$-to-$Y$ attention refers to using the tokens in $X$ to create queries and using the tokens in $Y$ to create keys and values. When $X=Y$, it is self-attention. The new attention score matrix $A$ is defined as
\begin{equation}
\begin{split}
    A = softmax\left(\frac{{X} W_{Q} ({Y} W_{K})^{T} R_{1} + {X} W_{Q} R_{2}}{\sqrt{d}}\right)
\end{split}
\label{eq:cross-attention}
\end{equation}
$W_Q$, $W_K$, $W_V$ are individual linear projections to generate the query, key and value embeddings. We note that in the implementation, the linear projections can be replaced by MLPs. $R_{1}$ and $R_{2}$ are two matrices encoding relative positions between tokens. $d$ is the embedding dimension of a single attention head.

To better explain \eqref{eq:cross-attention}, we describe this attention from the perspective of individual tokens.
We incorporate the sinusoidal functions to capture the relative position information,  which is illustrated as:
\begin{equation}
    q_{i} = 
    \begin{bmatrix}
        {x}_{i} W_{Q} \odot \cos{\left(p_{i}W_{1}+b_{1}\right)} \\
        {x}_{i} W_{Q} \odot \sin{\left(p_{i}W_{1}+b_{1}\right)} \\
        {x}_{i} W_{Q} \odot \cos{\left(p_{i}W_{2}+b_{2}\right)} \\
        {x}_{i} W_{Q} \odot \sin{\left(p_{i}W_{2}+b_{2}\right)} 
    \end{bmatrix}
\end{equation}

\begin{equation}
    k_{j} = 
    \begin{bmatrix}
        {y}_{j} W_{K} \odot \cos{\left(p_{j}W_{1}\right)} \\
        {y}_{j} W_{K} \odot \sin{\left(p_{j}W_{1}\right)} \\
        \cos{\left(p_{j}W_{2}\right)} \\
        \sin{\left(p_{j}W_{2}\right)} 
    \end{bmatrix}
\end{equation}
In the above equations, $q_{i}$ denotes a query originated from the token $x_{i}$ (i.e., the $i$-th row of $X$), and $k_{j}$ denotes a key originated from the token $y_{j}$ (i.e., the $j$-th row of $Y$). $p_i$ is the spatial position of the token $x_{i}$. $p_j$ is the spatial position of the token $y_{j}$.
$W_{1}$, $W_{2}$ are trainable weight matrices, and $b_{1}$, $b_{2}$ are trainable bias vectors inside the sine and cosine functions to generate different frequency components.
$\odot$ denotes the element-wise product of two tensors (vectors or matrices).

Thus, we derive each entry $a_{ij}$ of the attention score matrix $A$ from the scalar dot product of query vector $q_{i}$ and key vector $k_{j}$, which is given by:
\begin{equation}
\begin{split}
log(a_{ij}) &\propto sum\left(q_{i} \odot k_{j} \right) \\
&= sum \big( {x}_i W_Q \odot {y}_j W_K \odot \cos\left((p_i - p_j) W_1 + b_1 \right) \\
&\quad + {x}_i W_Q \odot \cos\left((p_i - p_j) W_2 + b_2 \right)\big)
\end{split}
\end{equation}

The output of an attention head has two terms: a term related to value (i.e., $YW_V$), and a term related to the relative positions. The $i$-th row of the output is calculated by
\begin{equation}
\label{eq:cross-attention-output}
% out_{i} = Linear\left((A_i X W_V)^T\right) + Linear\left( \sum\mathop{}_{\mkern-5mu j} a_{ij}\left(p_i - p_j\right) \right)
out_{i} = \text{Linear}\big(\left(A_i {Y} W_V\right)^T\big) + \text{Linear}\big( \sum_{j} a_{ij}\left(p_i - p_j\right) \big)
\end{equation}
where $A_i$ refers to the $i$-th row of the attention score matrix, \text{Linear} refers to a linear layer with trainable weight and bias. 
%Our implementation supports multi-head attention.  

%The new attention formula not only supports image self-attention but also enables shape self-attention and shape-to-image attention. Our implementation supports multi-head attention.  

\subsection{UNet-DeformSA}

\begin{figure}[htbp]
% Syntax: \includegraphics[trim={left bottom right top},clip]{filename}
\centerline{\includegraphics[width=\columnwidth]{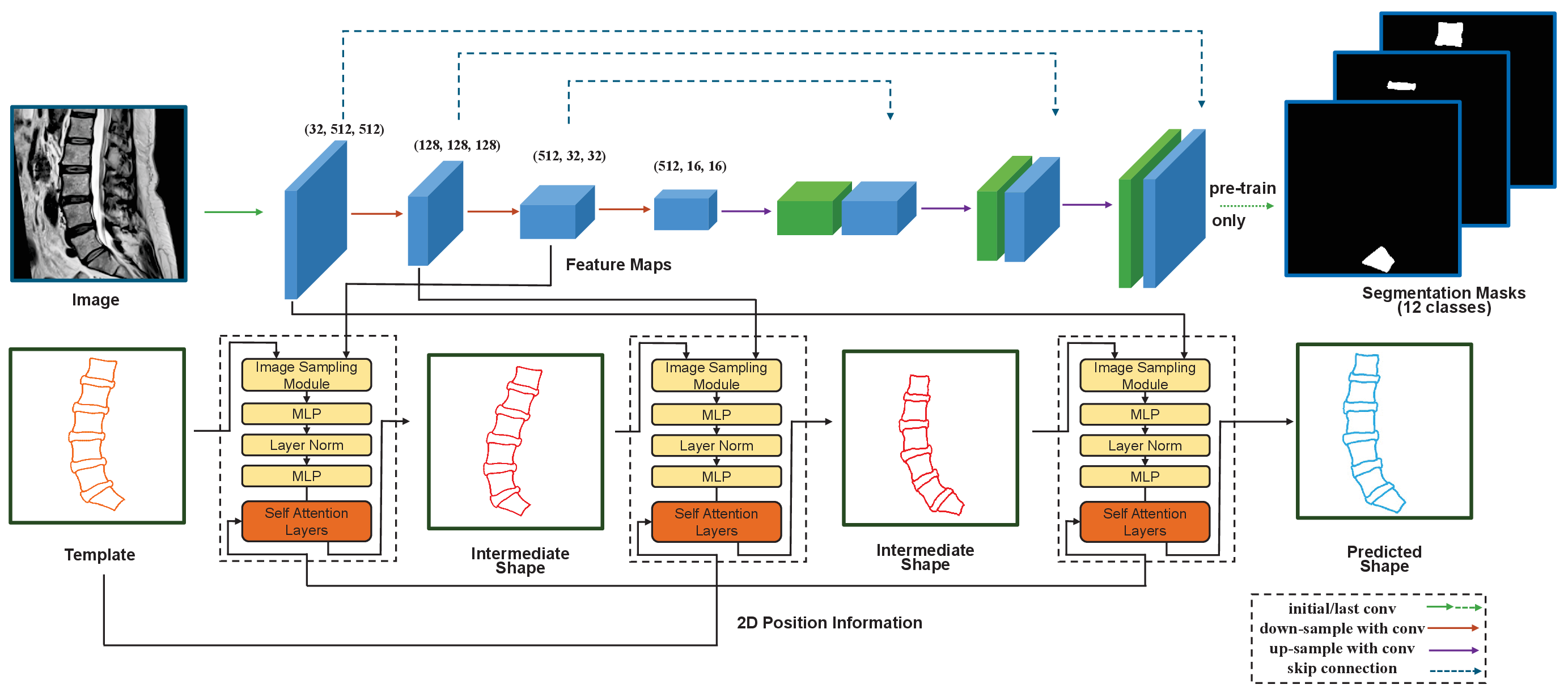}}
\caption{
Structure of UNet-DeformSA. The lumbar spine geometry is reconstructed gradually through three geometry-deformation modules. Each of the modules has shape self-attention layers.
}
\label{fig:unet-deformsa-model}
\end{figure}
% [Reminder] This position is better for placing the transdeformer model
\begin{figure*}[htbp]
% Syntax: \includegraphics[trim={left bottom right top},clip]{filename}
\centerline{\includegraphics[width=\textwidth]{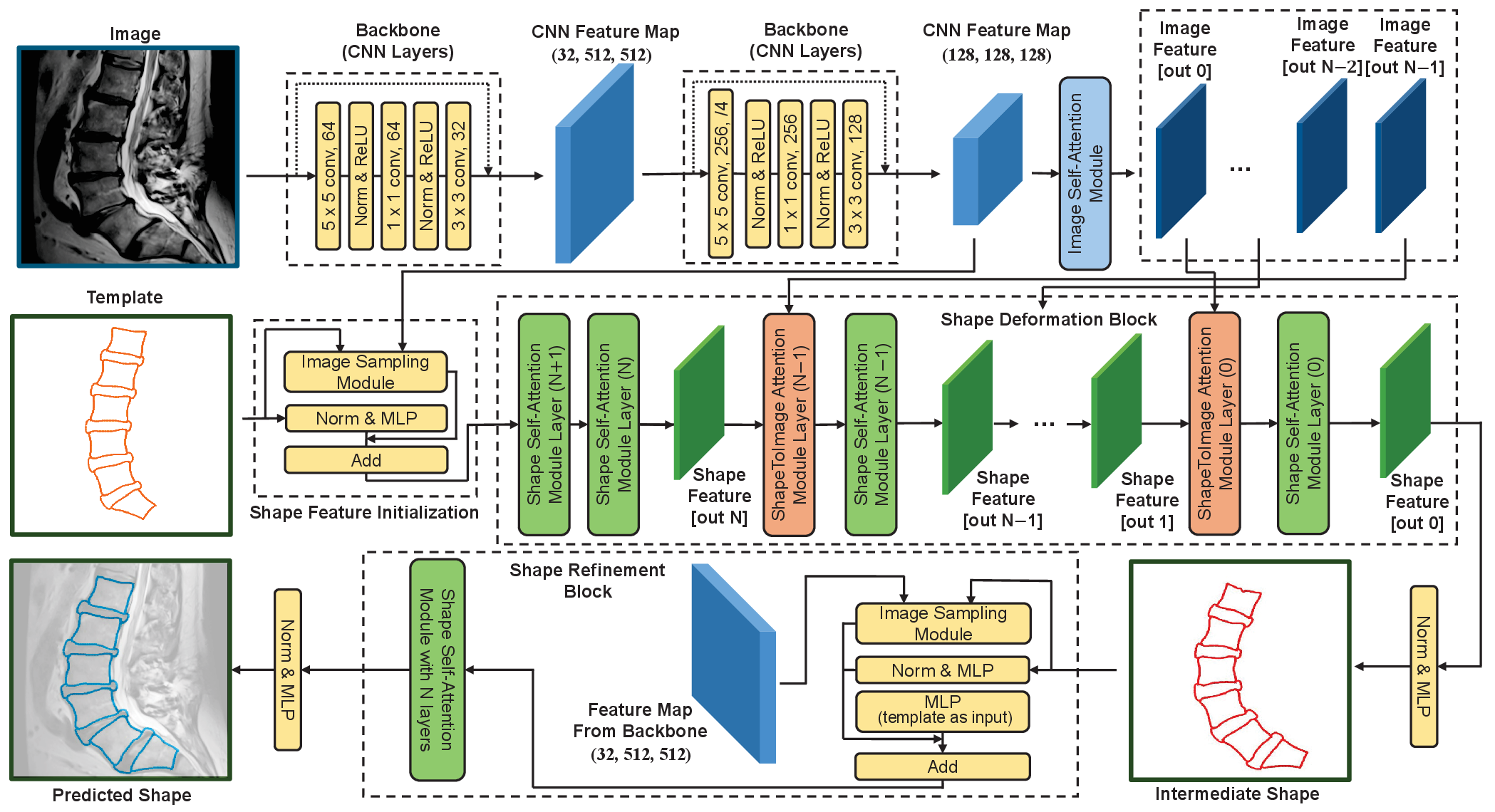}}
\caption{
Structure of TransDeformer.
}
\label{fig:net6}
\end{figure*}

The architecture of UNet-DeformSA is shown in Fig. \ref{fig:unet-deformsa-model}. The UNet backbone facilitates the learning of a hierarchical representation of a lumbar spine MR image (512$\times$512), which consists of 4 encoding layers that output feature maps at 4 spatial resolutions (i.e., 512$\times$512, 128$\times$128, 32$\times$32, and 16$\times$16), and 4 decoding layers with residual connections from the encoding layers. In our experiments, the UNet backbone is pretrained with paired images and segmentation masks, and then the backbone is frozen for feature extraction to serve the other parts of the network. The UNet-DeformSA model has three geometry-deformation modules, and each of which has an image sampling layer and shape self-attention layers. Each geometry-deformation module predicts the displacement vectors at the points of the template shape. In this way, the template shape is deformed gradually by the modules. The shape self-attention layers and the image sampling layer in UNet-DeformSA are the same as those in TransDeformer, and the details are provided in the next section.

%In geometry deformation module, we decompose the shape-specific deformation into three different stages: (1) In the first stage, the centroid of the predicted lumbar spine geometry (i.e., the arithmetic average of all points on the shape) is aligned with the ground truth. (2) In the second stage, the displacement vectors at each point of lumbar shape are predicted. (3) The third stage works similar as the second stage, in which an adjustment of displacement vectors is predicted. In brief, we align the shape’s centroid, the arithmetic average position of all points, with the ground truth, and predict minor displacement at each point of the shape. This enables each stage to specialize in different aspects of deformation, thereby enhancing the accuracy and robustness. The final output is the deformed geometry of the lumbar spine, represented by the connected contours of vertebrae and discs. Segmentation masks are then generated by converting the contours/polygons of vertebrae and discs to individual segmented regions.

\subsection{TransDeformer}
The architecture of the TransDeformer model is depicted in Fig. \ref{fig:net6}. Instead of using a UNet, TransDeformer has two groups of CNN layers serving as two feature extractors at two spatial resolutions (512$\times$512 and 128$\times$128). Given the input template shape, the image sampling module extracts image features at each point of the template shape from the feature maps at the lower resolution. Then, image self-attention, shape self-attention, and shape-to-image attention will be performed to cross-examine and fuse information gathered from individual image patches of the feature maps and points of the template. The final output of the shape self-attention module is used to predict the displacement vectors at the points of the template shape, and then the template is deformed to obtain an ``intermediate" shape that is further refined by using the higher resolution feature maps, and this refinement is done through a combination of an image sampling layer and a shape self-attention module that outputs adjustments to the ``intermediate" displacement vectors. The refined displacement vectors produce the final output shape. The major components of the TransDeformer model are explained with more details in the following sub-sections.

\subsubsection{Image Self-Attention (ISA) Module}
\begin{figure}[htbp]
% Syntax: \includegraphics[trim={left bottom right top},clip]{filename}
\centerline{\includegraphics[width=\columnwidth]{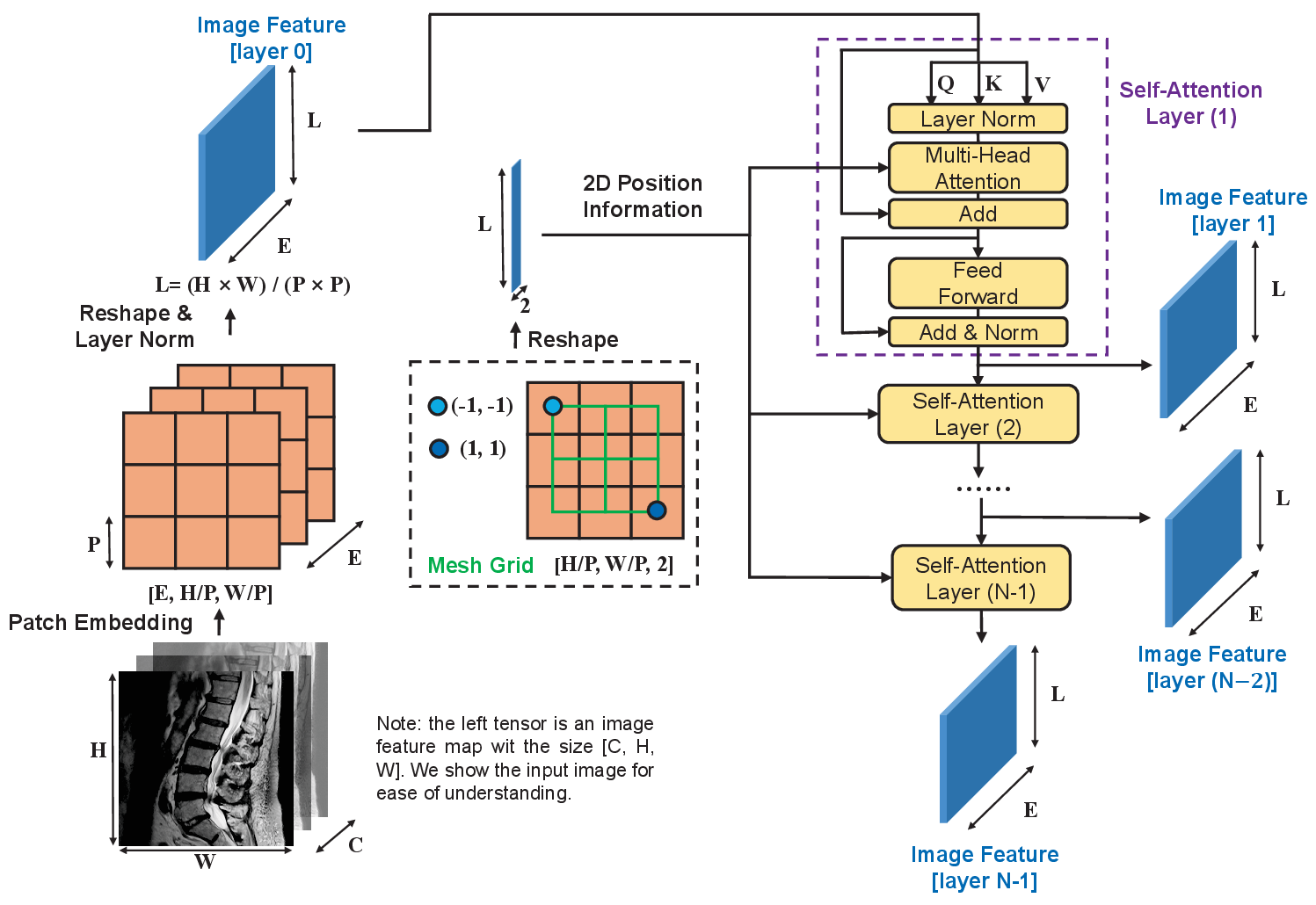}}
\caption{
The overview of the image self-attention (ISA) module.
}
\label{fig:isa}
\end{figure}

The architecture is depicted in Fig. \ref{fig:isa}.
The ISA module aims to produce the sequential image features with the aggregation of global context from the input CNN feature maps.
The CNN feature maps are divided into non-overlapping patches \cite{b24}, thereby enabling contextual dependencies across patches as opposed to individual pixels. 
%This boosts computational efficiency and renders the module resilient to disturbances on image pixels.
A patch is a token in the attention mechanism described in Section A.
A mesh grid, $G\in{\mathbb{R}^{\frac{H}{P} \times \frac{W}{P} \times 2}}$ is used for position embedding, where $P$ is the number of patches, $H$ and $W$ are spatial height and width of a patch. Each patch is centered on one of the mesh grid points. The mesh grid coordinates are normalized between -1 and 1.
%to bridge the gap between the tokenized image patches and their embedded coordinates by discretization on the domain of image.
%Let us denote the mesh grid as $G\in{\mathbb{R}^{\frac{H}{P} \times \frac{W}{P} \times 2}}$ where $P$ is the number of patches, $H$ and $W$ are spatial height and width.  
%Here, this mesh grid embodies a coordinate grid covering the $\frac{H}{P} \times \frac{W}{P}$ discrete domain of patches where each patch corresponds to its relative position as a tuple $G_{i,j}: (\frac{P\times i}{H}, \frac{P \times j}{W})$, ensuring a consistent mapping.
%and a feed-forward network where a series of image features are generated.
The ISA module has multiple self-attention layers to generate a sequence of feature matrices, enabling conceptual representations with increasingly long-range associate information. Each row of a feature matrix ${Y_I}\in{\mathbb{R}^{L\times E}}$ is a token, where $L$ is the number of patches/tokens and $E$ is the embedding/feature dimension.

\subsubsection{Image Sampling Module}
At each point of the template shape, an image feature vector is extracted by bilinear interpolation in the feature maps and then combined with a position embedding vector. A shape is described not only by a set of points but also by embeddings/features at those points. 

\subsubsection{Shape Self-Attention (SSA) Module}
\begin{figure}[htbp]
% Syntax: \includegraphics[trim={left bottom right top},clip]{filename}
\centerline{\includegraphics[width=\columnwidth]{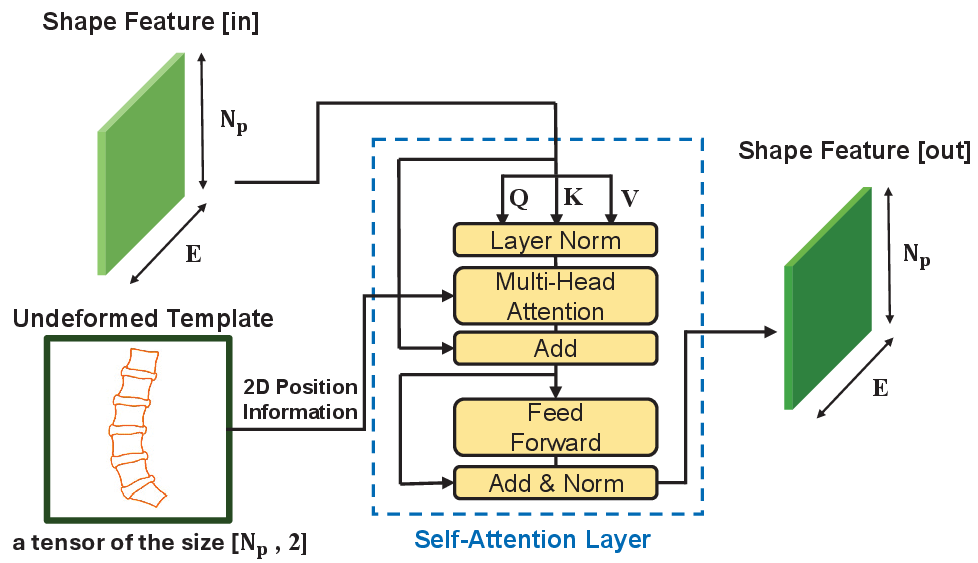}}
\caption{
The overview of the shape self-attention (SSA) module
}
\label{fig:ssa}
\end{figure}

The architecture is depicted in Fig. \ref{fig:ssa}.
The SSA module aims to learn contextual dependencies across the points of a shape, where individual points are associated  with embedding vectors that collectively form the shape feature matrix ${X_S}\in{\mathbb{R}^{N_{p}\times E}}$. $N_{p}$ is the number of points/tokens of the shape and $E$ is the embedding/feature dimension. A row in the feature matrix ${X_S}$ is a token in the attention mechanism described in Section A. For self-attention calculation, the position of a token is defined on the undeformed template with normalized coordinates between -1 and 1. The SSA module has multiple self-attention layers to generate a sequence of feature matrices.
%As edges function in a graph as connection, the initial template which records the original physical positions of coordinates is utilized as the position embedding to ensure contour consistency and act as the regularization mechanism. 
%After that, we incorporate the self-attention layers and a feed-forward network to refine the shape feature with long-range dependency captured.

\subsubsection{Shape-to-Image Attention (S2IA) Module}
\begin{figure}[htbp]
% Syntax: \includegraphics[trim={left bottom right top},clip]{filename}
\centerline{\includegraphics[width=\columnwidth]{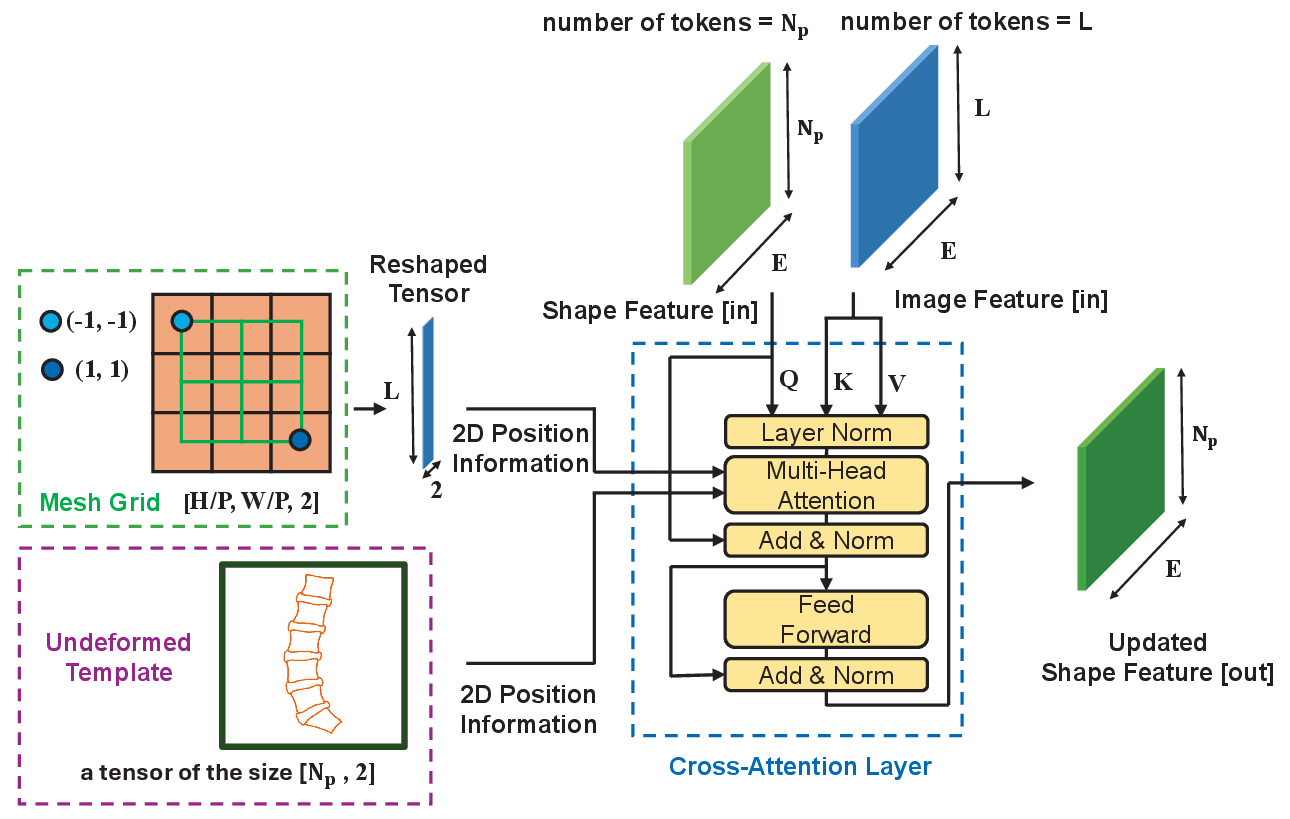}}
\caption{
The overview of the shape-to-image attention (S2IA) module
}
\label{fig:s2ia}
\end{figure}

The architecture is depicted in Fig. \ref{fig:s2ia}, serving as a bridge between the ISA and SSA modules. Given ${Y_I}$ from a layer of the ISA module and ${X_S}$ from a layer of the SSA module, a new shape feature matrix $\hat{X}_S\in{\mathbb{R}^{N_{p}\times E}}$ is generated by the S2IA module and becomes the input to the next shape self-attention layer. In the S2IA module, the position of a token in ${X_S}$ is defined in the image space to be compatible with the position of a token in ${Y_I}$.  ${X_S}$-to-${Y_I}$ attention follows the formula in Section A, and the output of ${X_S}$-to-${Y_I}$ attention is $\hat{X}_S$.
%We propose the S2IA module to facilitate the interaction between the shape tokens $\hat{S}\in{\mathbb{R}^{N_{p}\times E}}$ and image patch tokens $\hat{X}\in{\mathbb{R}^{L\times E}}$ where these tokens share the the same embedding dimension $E$. 
%Here, $N_p$ is the number of coordinates on the shape feature, and $L$ is the number of patches on the image feature.
%As previous approaches (Section \ref{sec:baseline-approaches}), the shape tokens initially integrate each coordinate and its neighbor pixels within the image feature via by sampling, thereby effectively filtering out irrelevant information.
%However, this interpolation is confined to the perception field, affecting the number of neighbor pixels covered by each coordinate.
%This constraint impedes connectivity between coordinates and the image feature, a critical aspect in medical imaging where different regions remain strong connection.
%Compared with sampling approach, this process aligns each coordinate with the entire image patches more comprehensively due to the extracted global context by cross attention and its dynamic weighting process.

In the first shape self-attention layer, each token/point integrates information from its coordinates, neighbor pixels in the image feature maps, and all the other tokens. If the template is initialized close to the true shape of the lumbar spine, then the information at a token could be sufficient for predicting a displacement vector at the token/point. If the template is initialized far away from the true shape, then the displacement vector at a token could not be accurately predicted only by this layer due to insufficient information. Therefore, cross-attention between shape tokens in ${X_S}$ and image tokens in ${Y_I}$ is used to gather information from relevant image patches that may be far away from the current template shape, i.e., fostering the long-range dependencies.

%To address this connection, we utilize the attention mechanism across the later shape and image tokens and foster the long-range dependencies between each individual coordinate feature $\hat{s}\in{\mathbb{R}^{1\times E}}$ and the entire patch sequence $\hat{X}$.

\subsection{Loss Function, Training and Inference Strategies}
\label{sec:loss-and-stage-ts}
To train the UNet-DeformSA model, the loss $\mathcal{L}$ combines a segmentation term $\mathcal{L}_\text{seg}$ and a geometry term $\mathcal{L}_\text{geom}$:
\begin{equation}
    \mathcal{L} = \mathcal{L}_\text{seg} + \mathcal{L}_\text{geom}
\label{eq:loss}
\end{equation}

To train the TransDeformer model, only $\mathcal{L}_\text{geom}$ is used because the model does not need image segmentation.
$\mathcal{L}_\text{seg}$ is a blend of Dice loss and area-weighted cross-entropy loss. $\mathcal{L}_\text{geom}$ measures the difference between the predicted shape and the ground-truth shape. 

%Let $\overline{M}$, $M$ be the predicted and ground truth segmentation masks respectively.
Let $\overline{M_{i}}$ and $M_{i}$ denote the predicted and the ground truth binary segmentation masks of the $i$-th class respectively, where $i\in{\left\{0, 1, ..., n \right\}}$ and class-0 is the background.
$w_{i}$ is a nonnegative scalar inversely proportional to the area of the $i$-th class, with the constraint $\sum_{i=1}^{n} w_{i} = 1$.
%$\odot$ denotes the element-wise product.
The segmentation quality term $\mathcal{L}_\text{seg}$ is formed as:
\begin{equation}
\begin{split}
    \mathcal{L}_\text{seg} = & -\sum_{i=1}^{n} \left[ \frac{2 \times \sum(\overline{M_{i}} \odot M_{i})}{\sum \overline{M_{i}} + \sum M_{i}} \right] \\
    & - \sum_{i=1}^{n} \left[ \sum(w_{i} M_{i} \odot \log(\overline{M_{i}})) \right]
\end{split}
\label{eq:seg-loss}
\end{equation}

Also, we proposed a three-stage training strategy to improve spatial accuracy on meshes and robustness to template initialization.  

\subsubsection{Training Stage 1}
In the first stage, we let the models predict centroids only.
Let $ \hat{C}$ denote the centroid of a predicted shape $\hat{S}$ (i.e., deformed template). 
Let $C$ denote the centroid of the corresponding ground truth shape $S$. Then the loss 
of geometry quality is formulated as:
\begin{equation}
\label{eq:geometry-s0-loss}
    \mathcal{L}_{\text{geom}}^{(1)} (\hat{S}, S) = \Vert \hat{C} - C \Vert^{2}_{2}
\end{equation}
The template is randomly placed in an image, and then the image and the template are fed to our models during training.

\subsubsection{Training Stage 2}
In the second stage, we let the models predict whole shapes. The loss measures the mean squared error (MSE) of the predicted shape $\hat{S}$, compared to the ground-truth shape $S$ (a flattened array of coordinates of all the points of the shape).
\begin{equation}
\label{eq:geometry-s2-loss}
    \mathcal{L}_{\text{geom}}^{(2)}(\hat{S}, S) = \frac{1}{N_p}\Vert \hat{S} - S \Vert^{2}_{2}
\end{equation}
The template is initialized close to the true shape in an image, and then the image and the template are fed to our models during training.

\subsubsection{Training Stage 3}
The loss in the third stage $\mathcal{L}_{\text{geom}}^{(3)} (\hat{S}, S)$ is the same as that in the second stage. The difference is that in the third stage, nonlinear transform is applied to the template before model training. The transformed template is initialized close to the true shape in an image, and then the image and the template are fed to our models during training.

Our models have intermediate shape outputs: UNet-DeformsA has three intermediate shape outputs, and TransDeformer has one intermediate shape output. Our loss function considers not only the final output but also the intermediate shape outputs. The loss of geometry quality at each stage is a combination of the loss terms of the intermediate and final shape outputs.

\begin{equation}
\label{eq:geometry-overview-loss}
    \mathcal{L}_{\text{geom}}^{(t)} = \sum_{m}{\mathcal{L}_{\text{geom}}^{(t)}(\hat{S}^{(m)}, S)}
\end{equation}
where $\hat{S}^{(m)}$ refers to an intermediate shape or the final shape output, and $t$ is stage index.

We used a two-stage inference strategy. In the first stage, the template is initialized in the input image (i.e., template initialization), and the centroid displacement of the template is predicted. In the second stage, the template is re-initialized at the predicted centroid, and then the displacement vectors at individual points of the template are predicted. In experiments, we tested two types of template initialization: random initialization or placing the template at the center of an image.

\subsection{Error Estimation Using a Modified TransDeformer}
\label{sec:bias-estimation}
A model for error estimation takes an image and a shape $\hat{S}$ as the inputs, and predicts the difference between the input shape $\hat{S}$ and the ground-truth shape $S$ of the object in the image. The shape input to the error estimation model could be the output from a shape-reconstruction model (e.g., UNet-DeformSA).

\begin{figure}[h]
% Syntax: \includegraphics[trim={left bottom right top},clip]{filename}
\centerline{\includegraphics[width=\columnwidth]{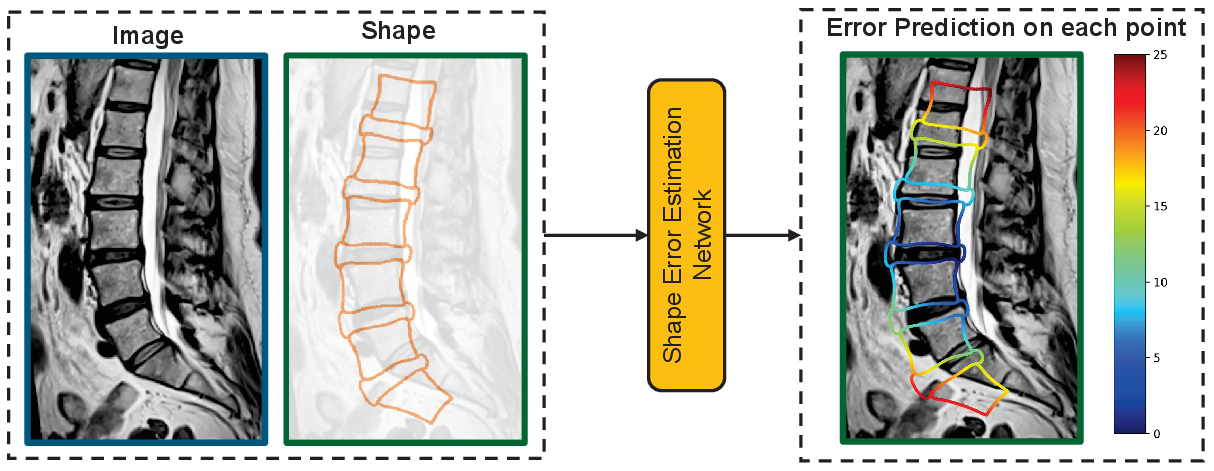}}
\caption{
The overview of the shape error estimation model (unit: mm).
}
\label{fig:shape-error-net}
\end{figure}

We built such an error estimation model (Fig. \ref{fig:shape-error-net}) by modifying the TransDeformer in the follow steps: (1) take a shape and an image as inputs, (2) remove/bypass the layers that predict the intermediate shape, and (3) output a nonnegative scalar at each point of the input shape. A scalar is an estimation of the distance between a point of the input shape and the corresponding point of the ground-truth shape. During model training, the input shape is random selected from the training dataset and then added with noises, and MSE loss is used.
%To demonstrate the estimated bias on the existing methods in Section \ref{sec:methods}, we utilized a transformer-based network.
%We employed the Pearson correlation coefficient between the estimated bias and the ground truth bias of the predicted contours to evaluate the bias.
%The larger correlation suggested that the bias was more precisely estimated by the network, thereby leading to larger bias between the predicted contour and ground truth.
%In our experiment, our model achieved the smaller correlation value, implying that our model is more reliable.
%\begin{figure}[htbp]

\section{Experiments}
\subsection{Dataset}
\begin{figure}[htbp]
% Syntax: \includegraphics[trim={left bottom right top},clip]{filename}
\centerline{\includegraphics[width=\columnwidth]{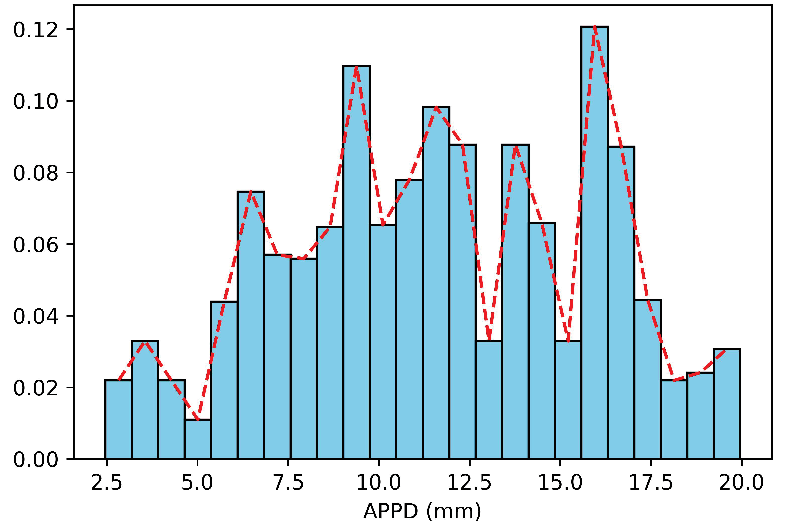}}
\caption{
Distribution of Average Point-to-Point Distance (APPD) between the template mesh and the ground truth meshes of the patients in the test set. APPD is defined in Section \ref{sec:D-metrics-and-results}.
}
\label{fig:appd-test-set}
\end{figure}

Our dataset \cite{b33} consists of de-identified lumbar spine MR images of 100 patients.
Three experts annotated and reviewed each mid-sagittal MR image to ensure accuracy and consistency.
Each mid-sagittal MR image was manually marked with boundaries and landmarks of the lumbar vertebrae and discs, which is guided by the established protocol \cite{b11}.
The images are resized to 512 $\times$ 512. Each image is accompanied with 12 segmentation masks, corresponding to the background, 6 vertebrae (named L1, L2, L3, L3, L4, L5, and S1), and 5 discs (named D1, D2, D3, D4, and D5).
We split the entire dataset into 70 training samples, 10 validation samples, and 20 test samples.
A special data augmentation method was applied to the original dataset, resulting in 7000 training samples, 250 validation samples, and 2500 test samples. The data augmentation method is described in detail in our technical report \cite{b33}. Our experiments were conducted on the augmented datasets that are publicly available \cite{b33}. Data diversity is illustrated in Fig. \ref{fig:appd-test-set} by comparing the template with the ground-truth meshes, for which the centroids are aligned.

\begin{table*}[ht]
\caption{Performance measured by APPD (lower is better)}
\label{tab:appd-overview}
\resizebox{\textwidth}{!}{
\centering
\begin{tabular}{ccccc|cccc|cccc|cccc|cccc}

\hline
\noalign{\vskip 2pt} % Adds space near the bottom hline

      & \multicolumn{4}{c|}{UNet-MLP}  & \multicolumn{4}{c|}{UNet-GCN}            & \multicolumn{4}{c|}{UNet-Disp} & \multicolumn{4}{c|}{UNet-DeformSA}               & \multicolumn{4}{c}{TransDeformer}                        \\
      & mean  & std   & max   & \multicolumn{1}{c|}{Q95}    & mean  & std   & max   & \multicolumn{1}{c|}{Q95}             & mean  & std   & max   & \multicolumn{1}{c|}{Q95}    & mean           & std   & max            & \multicolumn{1}{c|}{Q95}    & mean           & std   & max            & Q95             \\
\noalign{\vskip 2pt} % Adds space near the bottom hline

\hline
\noalign{\vskip 2pt} % Adds space near the bottom hline

L1      & 4.152 & 3.937 & 21.43 & 13.09 & 2.084 & 3.739 & 43.27 & 9.132 & 1.049 & 0.897 & 18.3  & 1.922 & \textbf{0.608}  & 0.223  & \textbf{2.278}  & \textbf{1.039}  & 0.647  & 0.325  & 2.937  & 1.319  \\
D1      & 2.869 & 2.877 & 23.09 & 8.823 & 2.151 & 4.587 & 47.75 & 10.62 & 1.257 & 1.336 & 18.39 & 2.592 & \textbf{0.653}  & 0.217  & \textbf{1.629}  & \textbf{1.106}  & 0.692  & 0.236  & 1.883  & 1.17   \\
L2      & 1.940 & 1.552 & 15.21 & 4.555 & 1.549 & 2.846 & 37.23 & 6.061 & 1.196 & 1.457 & 16.43 & 2.305 & \textbf{0.584}  & 0.216  & 1.392  & 1.047  & 0.589  & 0.211  & \textbf{1.297}  & \textbf{0.987}  \\
D2      & 1.667 & 1.816 & 20.18 & 4.000 & 1.019 & 1.161 & 16.67 & 2.672 & 1.139 & 1.117 & 15.39 & 2.236 & 0.681  & 0.291  & 3.993  & \textbf{1.164}  & \textbf{0.679}  & 0.344  & \textbf{2.884}  & 1.314  \\
L3      & 1.746 & 1.504 & 13.76 & 4.710 & 1.207 & 1.277 & 9.961 & 3.503 & 1.129 & 1.014 & 14.59 & 2.828 & \textbf{0.819}  & 0.452  & 4.538  & \textbf{1.799}  & 0.824  & 0.499  & \textbf{3.847}  & 1.801  \\
D3      & 1.586 & 0.870 & 7.584 & 3.401 & 1.118 & 0.789 & 10.18 & 2.827 & 1.072 & 0.495 & 8.664 & 2.023 & 0.905  & 0.482  & 3.191  & 2.079  & \textbf{0.876}  & 0.427  & \textbf{2.969}  & \textbf{1.847}  \\
L4      & 1.692 & 0.855 & 10.51 & 3.327 & 1.067 & 0.804 & 16.25 & 2.328 & 0.966 & 0.420 & 5.217 & 1.747 & 0.864  & 0.407  & 5.609  & 1.535  & \textbf{0.787}  & 0.296  & \textbf{2.455}  & \textbf{1.301}  \\
D4      & 2.375 & 1.789 & 23.62 & 5.178 & 1.359 & 1.315 & 26.73 & 2.701 & 1.165 & 0.570 & 4.697 & 2.327 & \textbf{0.983}  & 0.543  & 8.554  & 1.794  & 0.987  & 0.383  & \textbf{3.412}  & \textbf{1.723}  \\
L5      & 2.462 & 2.156 & 25.42 & 6.046 & 1.366 & 1.480 & 28.08 & 2.842 & 0.940 & 0.471 & 5.277 & 1.860 & 0.758  & 0.372  & 4.022  & 1.408  & \textbf{0.750}  & 0.292  & \textbf{2.712}  & \textbf{1.301}  \\
D5      & 2.554 & 2.855 & 34.47 & 6.436 & 1.291 & 1.119 & 21.38 & 2.435 & 1.134 & 0.524 & 5.356 & 2.103 & \textbf{0.759}  & 0.413  & 11.78  & \textbf{1.27}   & 0.785  & 0.315  & \textbf{2.406}  & 1.415  \\
S1      & 2.945 & 2.935 & 33.96 & 7.214 & 1.324 & 0.677 & 7.155 & 2.507 & 1.166 & 0.555 & 5.126 & 2.310 & 0.887  & 0.801  & 23.35  & 1.726  & \textbf{0.818}  & 0.365  & \textbf{3.201}  & \textbf{1.511}  \\
Whole   & 2.458 & 1.272 & 10.62 & 5.155 & 1.435 & 1.094 & 10.51 & 3.385 & 1.107 & 0.588 & 10.5  & 1.819 & 0.785  & 0.259  & 5.074  & 1.25   & \textbf{0.769}  & 0.212  & \textbf{1.836}  & \textbf{1.198}  \\
\hline
\multicolumn{21}{l}{For a patient, the average of the 11 APPD scores is used as a summary score; mean, std, max, Q95 of the summary scores are reported in the `Whole' row.}
\end{tabular}}
\end{table*}

\begin{table*}[ht]
\caption{Performance measured by Dice (higher is better)}
\label{tab:dsc-overview}
\resizebox{\textwidth}{!}{
\centering
\begin{tabular}{ccccc|cccc|cccc|cccc|cccc}

\hline
\noalign{\vskip 2pt} % Adds space near the bottom hline

      & \multicolumn{4}{c|}{UNet-MLP}  & \multicolumn{4}{c|}{UNet-GCN}            & \multicolumn{4}{c|}{UNet-Disp} & \multicolumn{4}{c|}{UNet-DeformSA}               & \multicolumn{4}{c}{TransDeformer}                        \\
      & mean  & std   & min   & \multicolumn{1}{c|}{Q5}    & mean  & std   & min   & \multicolumn{1}{c|}{Q5}             & mean  & std   & min   & \multicolumn{1}{c|}{Q5}    & mean           & std   & min            & \multicolumn{1}{c|}{Q5}    & mean           & std   & min            & Q5             \\
\noalign{\vskip 2pt} % Adds space near the bottom hline

\hline
\noalign{\vskip 2pt} % Adds space near the bottom hline

L1    & 0.875 & 0.153 & 0.022 & 0.538 & 0.942 & 0.114 & 0.002 & 0.692          & 0.968 & 0.038 & 0.181 & 0.945 & \textbf{0.980} & 0.008 & \textbf{0.936} & \textbf{0.962} & 0.978          & 0.010 & 0.927          & 0.955 \\
D1    & 0.846 & 0.182 & 0.001 & 0.409 & 0.903 & 0.178 & 0.000 & 0.393          & 0.921 & 0.133 & 0.000 & 0.855 & \textbf{0.964} & 0.016 & \textbf{0.880} & \textbf{0.930} & 0.963          & 0.017 & 0.870          & 0.925 \\
L2    & 0.949 & 0.070 & 0.064 & 0.832 & 0.956 & 0.095 & 0.037 & 0.824          & 0.964 & 0.038 & 0.516 & 0.933 & \textbf{0.979} & 0.010 & 0.937          & 0.959 & 0.979          & 0.009 & \textbf{0.942} & \textbf{0.962} \\
D2    & 0.937 & 0.094 & 0.024 & 0.776 & 0.956 & 0.049 & 0.370 & 0.870          & 0.934 & 0.098 & 0.000 & 0.844 & \textbf{0.963} & 0.022 & 0.684          & \textbf{0.938} & 0.962          & 0.024 & \textbf{0.754} & 0.925 \\
L3    & 0.952 & 0.068 & 0.521 & 0.830 & 0.962 & 0.053 & 0.615 & 0.887          & 0.964 & 0.031 & 0.640 & 0.883 & \textbf{0.969} & 0.022 & 0.799          & \textbf{0.923} & 0.969          & 0.024 & \textbf{0.840} & 0.917 \\
D3    & 0.940 & 0.054 & 0.469 & 0.817 & 0.949 & 0.050 & 0.478 & 0.826          & 0.942 & 0.042 & 0.235 & 0.865 & 0.950          & 0.035 & 0.660          & 0.875 & \textbf{0.951} & 0.032 & \textbf{0.680} & \textbf{0.880} \\
L4    & 0.962 & 0.034 & 0.383 & 0.898 & 0.970 & 0.037 & 0.228 & 0.932          & 0.970 & 0.014 & 0.801 & 0.945 & 0.970          & 0.014 & 0.835          & 0.946 & \textbf{0.972} & 0.011 & \textbf{0.895} & \textbf{0.950} \\
D4    & 0.918 & 0.085 & 0.129 & 0.768 & 0.945 & 0.065 & 0.025 & 0.877          & 0.942 & 0.030 & 0.585 & 0.884 & \textbf{0.952} & 0.023 & 0.742          & \textbf{0.911} & 0.950          & 0.024 & \textbf{0.762} & 0.903 \\
L5    & 0.953 & 0.059 & 0.255 & 0.879 & 0.966 & 0.042 & 0.169 & 0.938          & 0.970 & 0.012 & 0.849 & 0.945 & 0.974          & 0.011 & \textbf{0.892} & 0.951 & \textbf{0.975} & 0.010 & 0.876          & \textbf{0.956} \\
D5    & 0.918 & 0.093 & 0.155 & 0.740 & 0.958 & 0.040 & 0.307 & 0.930 & 0.943 & 0.027 & 0.557 & 0.900 & \textbf{0.962} & 0.017 & 0.673          & \textbf{0.933} & 0.960          & 0.019 & \textbf{0.877} & 0.920          \\
S1    & 0.928 & 0.085 & 0.182 & 0.791 & 0.966 & 0.026 & 0.506 & 0.931          & 0.960 & 0.021 & 0.805 & 0.915 & 0.968 & 0.026 & 0.397          & \textbf{0.935} & \textbf{0.968}          & 0.017 & \textbf{0.884} & 0.934 \\
Whole & 0.925 & 0.052 & 0.594 & 0.813 & 0.952 & 0.041 & 0.623 & 0.869          & 0.952 & 0.033 & 0.487 & 0.914 & \textbf{0.967} & 0.012 & 0.856          & 0.942 & 0.966          & 0.012 & \textbf{0.881} & \textbf{0.942} \\
\hline
\multicolumn{21}{l}{For a patient, the average of the 11 dice scores is used as a summary score; mean, std, min, Q5 of the summary scores are reported in the `Whole' row.}
\end{tabular}}
\end{table*}

\subsection{Comparison with Related Approaches}
\label{sec:baseline-approaches}
We adapted two related approaches \cite{b1, b20} for our application, which were  proposed for reconstructing organ geometries by deforming templates through neural networks. 

\subsubsection{UNet-GCN Model}
\label{sec:cnn-gcn-hybrid-model}
This model was originally proposed by Kong et al. \cite{b1} for reconstructing human heart geometries from 3D volumetric images. The model has a UNet backbone pretrained for binary segmentation, and the encoder of the UNet is used as the image feature extractor at different scales. Three GCN modules are used to deform a template sequentially by using image features sampled at the points of the template. For our application, we followed the original design as close as possible, and made necessary adjustment by changing the 3D UNet to a 2D UNet to handle 2D sagittal images of lumbar spine. The UNet-GCN model uses GCN layers to fuse information from neighbour points for displacement prediction. 
Our UNet-DeformSA model has shape self-attention modules to integrate information from different points that could be far away from each other (i.e., utilizing long-range dependency), which is the key difference from and advantage over the UNet-GCN model.
%a UNet-GCN framework integrating an image-encoderencoding backbone by UNet with three GCN modules that intake extracted features at different scales.  

%We implement the UNet-based model by incorporating GCN layers to regularize spatial interactions among contour points, as proposed by Fanwei et al. \cite{b1}. 
%This methodology mirrors the approach presented by Kong et al. \cite{b1}, aimed at reconstructing the anatomy of heart from 3D volumetric images.
%The CNN-GCN framework integrates an image-encoding backbone by UNet, along with three GCN modules that intake extracted features at different scales. 
%The backbone is pretrained with binary segmentation before bilinear interpolation, a resampling technique, is utilized to fuse the feature maps with intermediate contours.
%Additionally, we implement UNet-MLP model by simplifying this model in ablation study (Section \ref{sec:ablation-study}).

\subsubsection{UNet-Disp Model}
This model was originally proposed by Pak et al. \cite{b20} for creating simulation-ready heart valve geometries from 3D CT images. Compared to the UNet-GCN model, the UNet-Disp model also uses a UNet but in a different way: the decoder of the UNet predicts a displacement field that is used to deform a template. For our application, we followed the original design as close as possible, and made necessary adjustment by changing the 3D UNet to a 2D UNet that outputs a 2D displacement field. The UNet-Disp model does not use any  attention mechanisms. 
%We investigat the UNet-Displ model to manipulate a predefined template by a predicted space-deforming field on the regular grid \cite{b22}.
%The UNet-Displ framework inputs a lumbar spine image and predicts the control point displacements at uniform intervals, which are then applied to the reset of the image coordinates through a third order b-spline interpolation.
%The interpolated output represents the final diffeomorphic flow as a vector field \cite{b21}.
%The technique is proposed by Daniel et al. \cite{b20} for creating highly accurate, patient-specific cardiac geometries.

\subsubsection{UNet-MLP Model}
This model is a simplification of our UNet-DeformSA by replacing shape self-attention layers with a simple MLP in each geometry-deformation module.

\subsection{Implementation}
We used Pytorch ver. 2.0.0 to implement the overall frameworks, and used Pytorch Geometric ver. 2.3.0 to implement GCNs.
% We used Pytorch ver. 2.0.0 \cite{b30} to implement the overall frameworks, and used Pytorch Geometric ver. 2.3.0 \cite{b31} to implement GCNs.
%Our experiment involved the backbone \cite{b33} that integrated CNN and transformer branches with balanced weights across varying approaches.
%The CNN branch adopted the UNet architecture, consisting of 4 encoding layers with ResNet \cite{b32} at different resolutions (e.g., 512, 128, 32, 16) and 4 decoding layers concatenated by the residual connections from encoded features.
%The transformer branch followed the basic structure of the Vision Transformer, yet adjusted with different scales under the instruction \cite{b33}.
%This branch contained 4 transformer blocks, each was composed of 2 stacked attention layers with 16 heads, 512 embedding dimensions, and distinct patch scale (e.g., 16, 16, 4, 1) depending on its depth.
%The input to the GCN layers was the backbone's features sampled by coordinate-specific contours and an adjacency matrix of vertex connection.
%We investigated varying kernels such as ResGatedGraphConv \cite{b34}, ChebConv \cite{b35}, GATConv \cite{b36}, and TransformerConv \cite{b37} and a set of depths, and ultimately settled on the GCN block with 4 layers using the ResGatedGraphConv kernel.
%Similarly, following the instruction \cite{b20}, we explored the hyperparameter trade-off on the unitless scaling factor $\sigma$ within the various UNet-Displ models and setup $\sigma$ as 1.
%Furthermore, we setup the proposed TransDeformer model with a collection of ablation study in the later Section \ref{sec:ablation-study}.
We used the Adam optimizer 
%\cite{b25} 
with a fixed learning rate of 1e-4 and a batch size of 20. The training of each model completed within 24 hours on a single NVIDIA RTX A6000 with 48GB VRAM.

\subsection{Metrics and Results of Geometry Reconstruction}
\label{sec:D-metrics-and-results}
To measure the accuracy of a reconstructed lumbar object (a vertebra or a disc), we assessed two key metrics: (1) Average Point-to-Point Distance (APPD) between a predicted shape and the corresponding ground-truth shape and (2) Dice Similarity Coefficient (Dice), measuring the overlap between areas enclosed by predicted and ground-truth shapes. For each metric, mean, standard deviation (std), the worst case, the $95$th or $5$th percentile among the test samples are calculated. 

\begin{figure*}[htbp]
% Syntax: \includegraphics[trim={left bottom right top},clip]{filename}
\centering
\includegraphics[width=0.8\textwidth]{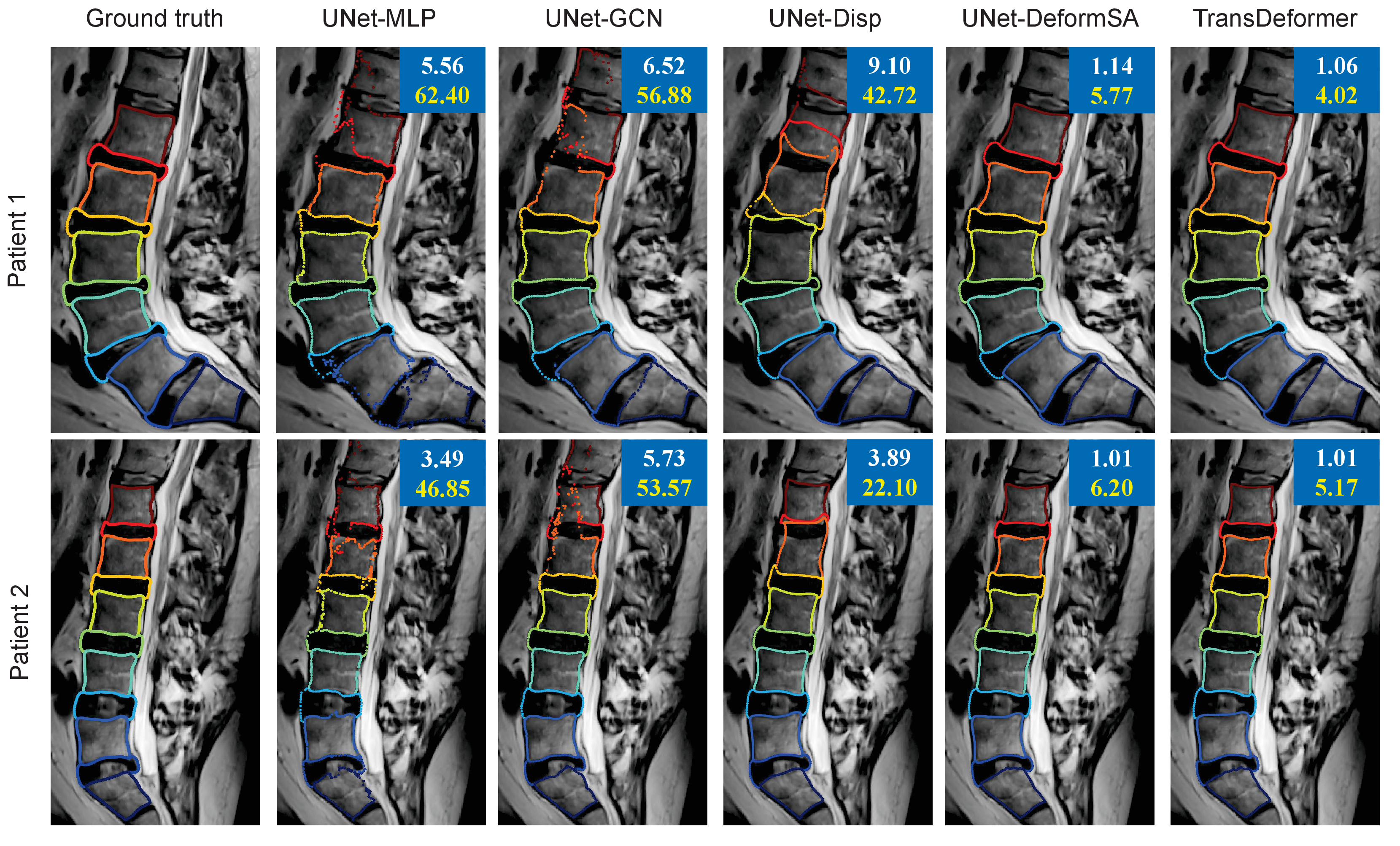}
% \centerline{\includegraphics[trim={10cm 3cm 10cm 5cm}, clip, width=\columnwidth]{figures/result-1.PNG}}
\caption{
Outputs of the models on two challenging cases. Different colors indicate different lumbar components. On the upper-right corner, the white number shows the average point-to-point distance error (mm), and the yellow number shows the maximum point-to-point distance error (mm).
}
\label{fig:abnormal-cases}
\end{figure*}

The results are reported in Table \ref{tab:appd-overview} and Table \ref{tab:dsc-overview}, showing our models outperform the other models. Examples are shown in Fig. \ref{fig:abnormal-cases}. Template initialization is done by placing the template at the center of an image.

%We also assessed our approach by DSC, which evaluated the performance on segmentation derived from the predicted contours.
%Our technique surpassed the other existing methods (Table. \ref{tab:net6-dsc}) since the segmentation enclosed by the generated contours remained accurate and consistent.
%As the minimal value reflected how server the segmentation contained artifacts even with extreme cases, the highest minimal value of DSC indicated that our model remarkably eliminated non-target areas.
%Whereas other methods demonstrated remarkable performance with segmentation included as pretrain, our approach forewent this additional supervision but managed to achieve superior quality with fewer artifacts.
%This implied that coordinate-specific contour remained informative and efficient in memory.

\subsection{Robustness to Template Initialization}
We conducted additional experiments to evaluate a model's robustness with respect to the initial location of the template. The template is placed at the center of an input image, and then a random shift within a circle is applied to the template before feeding it to a model. As the radius of the circle increases, a more robust model has a less reduction in accuracy. The results are reported in Table \ref{tab:robustness-to-template}, showing that our models are more robust than the other models.

\begin{table}[htbp]
\caption{Robustness to Template Initialization}
\label{tab:robustness-to-template}
\centering
%\setlength{\tabcolsep}{3pt}
%\begin{tabular}{
%|>{\centering\arraybackslash}p{0.22\columnwidth}
%*{5}{|>{\centering\arraybackslash}p{0.08\columnwidth}}|
%}
\begin{tabular}{cccccc}

\hline
\noalign{\vskip 2pt} % Adds space near the bottom hline

Radius (pixel) &  0     &  10    &  20     &  30     &  40     \\
\noalign{\vskip 2pt} % Adds space near the bottom hline
\hline
\noalign{\vskip 2pt} % Adds space near the bottom hline
UNet-MLP      & 2.458 & 2.498 & 2.760  & 4.581  & 8.743  \\
UNet-GCN      & 1.435 & 1.472 & 1.939  & 3.654  & 5.860  \\
UNet-Disp     & 1.107 & 6.548 & 13.19  & 19.52  & 25.67  \\
UNet-DeformSA & 0.784 & 0.785 & 0.798  & 0.879  & 1.155  \\
TransDeformer & \textbf{0.769} & \textbf{0.769} & \textbf{0.769}  & \textbf{0.769}  & \textbf{0.820}  \\
\hline
\multicolumn{6}{l}{Note: APPD (lower is better) is used in the robustness study}
\end{tabular}
\end{table}

\subsection{Performance of the Shape Error Estimation Model}
We used the shape error estimation model to estimate the errors of the outputs from each of the four models for geometry reconstruction, and computed the correlation coefficient between the estimated errors and the true errors. The result is reported in Table \ref{tab:shape-error-net-corr}.

\begin{table}[htbp]
\caption{Correlation ($\rho$) by the shape error estimation model}
\label{tab:shape-error-net-corr}
\centering
\setlength{\tabcolsep}{2.8pt}
\begin{tabular}{cccccc}
\hline
\noalign{\vskip 2pt} % Adds space near the bottom hline
            & UNet-MLP & UNet-GCN & UNet-Disp & UNet-DeformSA & TransDeformer \\
\noalign{\vskip 2pt} % Adds space near the bottom hline
\hline
\noalign{\vskip 2pt} % Adds space near the bottom hline
$\rho1$ & 93.55\%  & 73.55\% & 95.63\%   & 94.46\%       & 92.54\%      \\
\hline
\noalign{\vskip 2pt} % Adds space near the bottom hline
$\rho2$ & 96.08\%  & 94.86\% & 88.98\%   & 88.33\%       & 89.89\%      \\
\hline
\multicolumn{6}{l}{Note: $\rho1$ is Pearson Correlation; $\rho2$ is Spearman Correlation.}
\end{tabular}
\end{table}

\subsection{Medical Parameter Analysis}
\subsubsection{Medical Parameter Definitions}
\begin{figure}[htbp]
% Syntax: \includegraphics[trim={left bottom right top},clip]{filename}
\centerline{\includegraphics[width=\columnwidth]{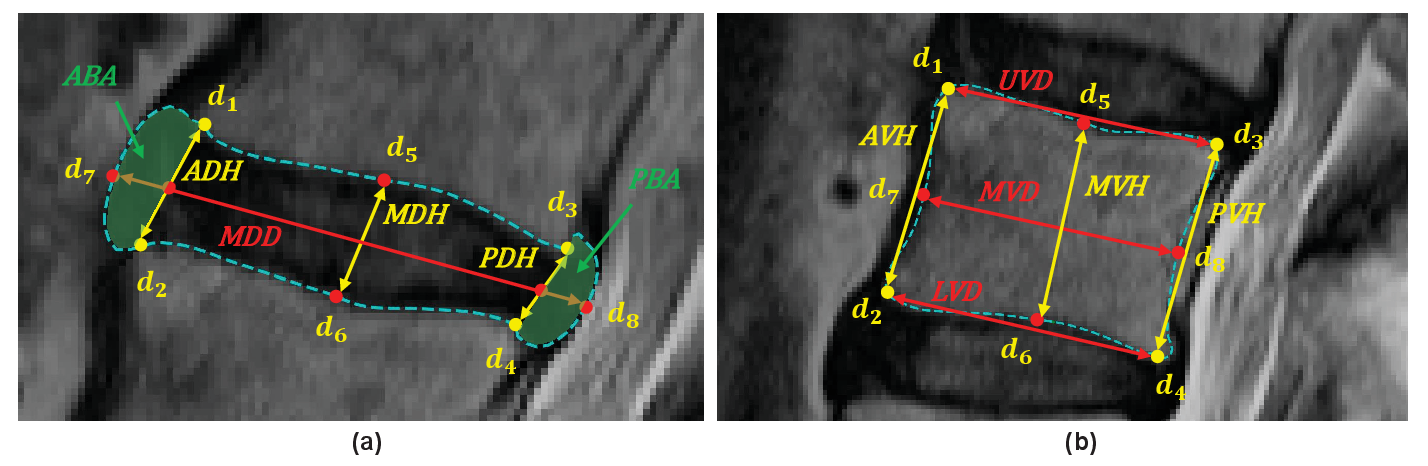}}
\caption{
Definitions of the medical parameters.
Corner vertices ($d_1$, $d_2$, $d_3$, and $d_4$), laying at the intersections between a vertebra and disc, are identified as landmarks.
Midpoints ($d_5$, $d_6$, $d_7$, and $d_8$) on edges are also identified.
For vertebrae, we introduce three main parameters, including upper, lower and middle vertebra diameters (UVD, LVD, MVD).
Additionally, we measure the height of anterior, posterior and middle vertebra (AVH, PVH, MVH) as well as the mean vertebra height (VHMean), a ratio of the vertebra's area to its diameter, to comprehensively depict the vertebra in cause of its abrasion.
Similar to vertebrae, we introduce three main parameters including anterior, posterior and middle disc heights (ADH, PHD, MDH) to measure disc dimensions.
The middle diameter of a disc (MDD) is defined as the line extending from the midpoints of landmarks and intersecting with the disc's left and right edges.
The mean disc height (DHMean) is calculated as the ratio of the disc's area to this diameter.
Moreover, relative anterior and posterior disc bulging areas (rABA, rPBA), ratios of the bulging areas to the averaged disc area, are introduced as critical metrics in evaluating disc degeneration.
}
\label{fig:medical-parameters}
\end{figure}

The template-deformation approach enables consistent definitions and measurements of medical parameters of the lumbar spine. Once a parameter definition is finalized on the template, the parameter of any patient's shape can be straightforwardly measured.
%Quantitatively, the coordinate-specific contour provided intuitive metrics for directly assessing key characteristics of the spinal components.
%This geometry representation maintained the point correspondence across different patients, thereby allowing for the consistent evaluation without any post-processing.
Following the previous work \cite{b11}, we included 15 medical parameters for disc degeneration assessment, which are explained in Fig. \ref{fig:medical-parameters}.
%, using comprehensive definitions of heights, diameters, and signal intensities of vertebrate and discs.
%The overall parameters are displayed in Fig. \ref{fig:medical-parameters}. The quantitative analysis is  decipted in Table. \ref{tab:net6-medical-parameters}.

%Here, we identified $4$ corner vertices (e.g., $v_1$ to $v_4$ or $d_1$ to $d_4$) laying at the intersections between a vertebra and disc as landmarks.
%Subsequently, we calculated $4$ midpoints (e.g., $v_5$ to $v_8$ or $d_5$ to $d_8$) on edges delineated by these landmarks to further assess the diameter and height of each vertebra and disc.
%For vertebrae, we introduced parameters such as upper, lower and middle vertebra diameters (UVD, LVD, MVD) to quantify vertebra dimensions on diameters.
%Additionally, we employed the height of anterior, posterior and middle vertebra (AVH, PVH, MVH) as well as the mean vertebra height (VHMean), a ratio of the vertebra's area to its diameter, to comprehensively depict the vertebra in cause of its abrasion.
%Similar to vertebrae, we introduced parameters like anterior, posterior and middle disc heights (ADH, PHD, MDH) to measure disc dimensions.
%Yet, the middle diameter of a disc (MDD) was defined as the line extending from the midpoints of landmarks to the intersections with the disc's left and right edges.
%Then, the mean disc height (DHMean) was calculated as the ratio of the disc's area to this diameter.
%Moreover, relative anterior and posterior disc bulding areas (rABA, rPBA), ratios of the bulging areas to the averaged disc area, wheere introduced as critical metrics in evaluating %disc degeneration.

\subsubsection{Parameter Measurement}
To evaluate the accuracy of the measured medical parameters using TransDeformer, we calculated the relative error of each parameter of a patient in the test set. Metrics are mean, standard deviation (std), max, and the $95$th percentile of the errors across the patients. Since the reconstructed geometries can be ranked by the errors estimated by the shape error estimation model, these metrics can be evaluated by using $\alpha$\% of the geometries with the lowest (estimated) errors, where $\alpha$ could be 20, 60, or 100. 
The results are reported in Table \ref{tab:transdeformer-medical-parameter}.

%, averaging these errors as our metric.
%For clear visualization, we expressed these relative errors as percentage (\%) in Table. \ref{tab:net6-medical-parameters}.
%Additionally, we utilized the estimated bias in Section \ref{sec:bias-estimation} to exclude cases of high bias for accurate evaluation.
%Also, we categorized tiered percentages (e.g., 100, 60, 20) where each percentage signified the proportion of cases with lower estimated bias relative to the whole test cases.

\begin{table*}[htbp]
\caption{Relative Error of Medical Parameter Measurement (lower is better)}
\label{tab:transdeformer-medical-parameter}
\resizebox{\textwidth}{!}{
\centering
\begin{tabular}{ccccc|cccc|cccc}

\hline
\noalign{\vskip 2pt} % Adds space near the bottom hline

\multirow{2}{*}{$\alpha$\%} & \multicolumn{4}{c|}{100\%}           & \multicolumn{4}{c|}{60\%}            & \multicolumn{4}{c}{20\%}            \\
                  & mean   & std    & max     & \multicolumn{1}{c|}{Q95}     & mean   & std    & max     & \multicolumn{1}{c|}{Q95}     & mean   & std    & max     & Q95     \\
\noalign{\vskip 2pt} % Adds space near the bottom hline

\hline
\noalign{\vskip 2pt} % Adds space near the bottom hline

UVD               & 1.75\% & 0.83\% & 6.44\%  & 3.61\%  & 1.49\% & 0.52\% & 4.24\%  & 2.49\%  & 1.45\% & 0.50\% & 3.67\%  & 2.35\%  \\
LVD               & 2.23\% & 0.91\% & 6.82\%  & 3.89\%  & 1.95\% & 0.74\% & 5.35\%  & 3.28\%  & 1.77\% & 0.65\% & 4.02\%  & 2.96\%  \\
MVD               & 2.44\% & 1.34\% & 7.65\%  & 5.09\%  & 2.02\% & 1.04\% & 7.08\%  & 4.02\%  & 1.76\% & 0.80\% & 4.31\%  & 3.61\%  \\
AVH               & 2.10\% & 1.08\% & 8.87\%  & 4.15\%  & 1.80\% & 0.85\% & 6.67\%  & 3.26\%  & 1.56\% & 0.66\% & 3.84\%  & 2.78\%  \\
PVH               & 2.24\% & 0.90\% & 6.64\%  & 3.93\%  & 1.98\% & 0.74\% & 6.08\%  & 3.32\%  & 1.89\% & 0.71\% & 4.57\%  & 3.20\%  \\
MVH               & 2.06\% & 1.33\% & 12.89\% & 4.58\%  & 1.59\% & 0.81\% & 5.33\%  & 3.14\%  & 1.33\% & 0.68\% & 3.78\%  & 2.72\%  \\
VHMean           & 1.87\% & 1.08\% & 8.91\%  & 4.08\%  & 1.43\% & 0.55\% & 3.52\%  & 2.42\%  & 1.26\% & 0.47\% & 2.76\%  & 2.12\%  \\
XY                & 1.21\% & 0.72\% & 7.37\%  & 2.55\%  & 1.04\% & 0.52\% & 3.34\%  & 2.02\%  & 0.98\% & 0.53\% & 2.96\%  & 1.99\%  \\
ADH               & 4.48\% & 2.20\% & 15.55\% & 8.95\%  & 3.68\% & 1.58\% & 11.49\% & 6.59\%  & 3.10\% & 1.21\% & 6.99\%  & 5.46\%  \\
PDH               & 7.23\% & 3.68\% & 28.22\% & 14.13\% & 6.42\% & 2.98\% & 17.27\% & 11.94\% & 6.43\% & 2.97\% & 16.42\% & 11.59\% \\
MDH               & 4.73\% & 2.82\% & 23.24\% & 10.21\% & 4.05\% & 2.17\% & 12.35\% & 8.29\%  & 3.67\% & 2.00\% & 11.29\% & 7.66\%  \\
DHMean           & 3.48\% & 1.94\% & 16.16\% & 6.88\%  & 2.84\% & 1.34\% & 8.14\%  & 5.36\%  & 2.60\% & 1.18\% & 6.37\%  & 4.84\%  \\
rABA               & 1.81\% & 0.86\% & 7.80\%  & 3.42\%  & 1.58\% & 0.58\% & 3.63\%  & 2.65\%  & 1.43\% & 0.48\% & 3.24\%  & 2.22\%  \\
rPBA               & 0.99\% & 0.44\% & 2.98\%  & 1.83\%  & 0.95\% & 0.42\% & 2.85\%  & 1.75\%  & 0.90\% & 0.37\% & 2.44\%  & 1.57\%  \\
disc area         & 3.59\% & 1.79\% & 12.69\% & 6.92\%  & 2.94\% & 1.21\% & 7.15\%  & 4.99\%  & 2.78\% & 1.08\% & 6.08\%  & 4.77\% \\
\hline
%\multicolumn{5}{l}{PPD: point-to-point distance}
\end{tabular}}
\end{table*}

%Here, most medical parameters maintained a relative error ratio of less than 2.5\%, effectively reflecting the characteristics of the lumbar spine.
%Also, as the percentage of cases with lower estimated bias increased, the relative error ratio correspondingly decreased as expected, due to the exclusion of 'hard' samples.

\subsection{Hyperparameter Tuning on Validation Set}
\label{sec:hyperparameter-study}
%Broadly, different hyperparameters can lead to vastly different predictions of the contour.
For the TransDeformer model, we investigated the effects of patch size $P$ in the ISA module and the number of attention layers in the ISA and SSA modules, and the results on the validation set are reported in Tables \ref{tab:as-net6-patchsize} and \ref{tab:as-net6-n-layers}. Based on the validation performance of the model, patch size of 4 and 2 layers of attention are used in the main experiments.

For the UNet-DeformSA model, we investigated the effect of the number of attention layers in the SSA module, and the results on the validation set are reported in Table \ref{tab:net1-n-layers}. 
Based on this results, 2 layers of attention are used in the main experiments. For the UNet-GCN model, we evaluated different types of GCN layers (Table \ref{tab:gcn-kernel-type}) and chose ResGatedGraphConv \cite{b34} based on the results on the validation set.
For the UNet-Disp model, we explored the hyperparameter $\sigma$ (Table \ref{tab:unetdp-sigma}) and set it to 1 based on the results on the validation set. The backbone UNet structure is the same for the three models.

%, both of which greatly contributed to contour's quality.
%The first observation highlighted how patch size influenced the accuracy of the deformed geometry.
%In the ISA module, a patch, seen as a token of the sequential image feature, captured local information within its non-overlapping window from others.
%Larger patches tended to grasp boarder and more abstract features but would overlook fine-grained details \cite{b38} whereas smaller patches increased the token count fed into the transformer, leading to longer sequences.
%This increment on the tokens boosted the computational load and called for additional parameters, potentially rendering the model more vulnerable to overfitting, particularly when data is insufficient.
%In Table. \ref{tab:as-net6-patchsize}, the patch size of 4 achieved optimal performance and maintained feasible GPU memory usage.

\begin{table}[ht]
\caption{Hyperparameter Tuning on Patch Size}
\label{tab:as-net6-patchsize}
\centering
\begin{tabular}{ccccc}
\hline
\noalign{\vskip 2pt} % Adds space near the bottom hline

\multirow{2}{*}{patch   size} & \multicolumn{4}{c}{APPD (mm)}  \\
                              & mean  & std   & max   & Q95   \\
\noalign{\vskip 2pt} % Adds space near the bottom hline

\hline
\noalign{\vskip 2pt} % Adds space near the bottom hline
2                             & 0.574 & 0.127 & 1.001 & 0.795 \\
4                             & \textbf{0.569} & 0.121 & \textbf{0.951} & \textbf{0.787} \\
8                             & 0.612 & 0.170 & 1.283 & 0.896 \\
16                            & 0.605 & 0.154 & 1.052 & 0.886\\
\hline
\multicolumn{5}{l}{Note: TransDeformer results (lower is better)}
\end{tabular}
\end{table}

%The second observation came from the number of identical layers $N$, also known as the depth, which played a critical role in the performance.
%A deeper transformer facilitates boarder information propagating across the entire sequence \cite{b39}, leading to richer abstractions on multiple levels but increased computational expense.
%In Table. \ref{tab:as-net6-n-layers}, $N=2$ offered the best performance on the maximum and $95$th percentile value of the contour's quality, measured by MSE on the coordinate-specific contour, with negligible differences on the average value.
%Also, this framework with the smaller depth considerably reduced memory and time cost.
%In summary, the TransDeformer model was implemented with 2 identical layers of the transformer where the attention layer were applied with 16 heads, 512 embedding dimensions and the patch size of 4.

%Additionally, we implemented another basic deformation-base model, UNet-style by simplifing the GCN layers in the CNN-GCN hybrid model with a set of multilayer perceptrons (MLPs).

%tested the basic deformation-based model with the UNet-style architecture, an advanced UNet which integrates weighted transformer layers, leveraging segmentation as supplementary guidance during the pretrain stage.
%A set of multilayer perceptrons (MLPs) utilize these sampled features to precisely predict the structured mesh of the lumbar spine through template deformation.

\begin{table}[ht]
\caption{Hyperparameter Tuning on the number of attention layers}
\label{tab:as-net6-n-layers}
\centering
\begin{tabular}{ccccc}
\hline
\noalign{\vskip 2pt} % Adds space near the bottom hline
\multirow{2}{*}{N} & \multicolumn{4}{c}{APPD (mm)}  \\
                   & mean  & std   & max   & Q95   \\
\noalign{\vskip 2pt} % Adds space near the bottom hline

\hline
\noalign{\vskip 2pt} % Adds space near the bottom hline
2                  & 0.569 & 0.121 & \textbf{0.951} & \textbf{0.787} \\
4                  & 0.581 & 0.137 & 0.981 & 0.851 \\
6                  & 0.593 & 0.150 & 1.017 & 0.851 \\
8                  & \textbf{0.562} & 0.140 & 1.023 & 0.821\\
\hline
\multicolumn{5}{l}{Note: TransDeformer results (lower is better)}
\end{tabular}
\end{table}

\begin{table}[ht]
\caption{Hyperparameter Tuning on the number of attention layers}
\label{tab:net1-n-layers}
\centering
\begin{tabular}{ccccc}
\hline
\noalign{\vskip 2pt} % Adds space near the bottom hline
\multirow{2}{*}{N} & \multicolumn{4}{c}{APPD (mm)}  \\
                   & mean  & std   & max   & Q95   \\
\noalign{\vskip 2pt} % Adds space near the bottom hline

\hline
\noalign{\vskip 2pt} % Adds space near the bottom hline
2       & 0.598     & 0.133     & \textbf{0.923}     & \textbf{0.820} \\
4       & 0.605     & 0.148     & 0.967     & 0.867 \\
6       & 0.595     & 0.150     & 0.969     & 0.849 \\
8       & \textbf{0.586}     & 0.144     & 0.948     & 0.837 \\
\hline
\multicolumn{5}{l}{Note: UNet-DeformSA results (lower is better)}
\end{tabular}
\end{table}

\begin{table}[ht]
\caption{Hyperparameter Tuning on the type of GCN layers}
\label{tab:gcn-kernel-type}
\centering
\begin{tabular}{ccccc}
\hline
\noalign{\vskip 2pt} % Adds space near the bottom hline
\multirow{2}{*}{GCN layer type} & \multicolumn{4}{c}{APPD (mm)}  \\
                   & mean  & std   & max   & Q95   \\
\noalign{\vskip 2pt} % Adds space near the bottom hline

\hline
\noalign{\vskip 2pt} % Adds space near the bottom hline
ResGatedGraphConv  & \textbf{0.779}  & 0.953  & 12.276          & \textbf{1.201} \\
ChebConv           & 0.936           & 1.003  & \textbf{11.913} & 2.104          \\
GATConv            & 2.191           & 2.993  & 15.890          & 9.587          \\
TransformerConv    & 1.595           & 2.395  & 14.250          & 7.154          \\
\hline
\multicolumn{5}{l}{Note: UNet-GCN results (lower is better)}
\end{tabular}
\end{table}

\begin{table}[ht]
\caption{Hyperparameter Tuning on the scaling factor $\sigma$}
\label{tab:unetdp-sigma}
\centering
\begin{tabular}{ccccc}
\hline
\noalign{\vskip 2pt} % Adds space near the bottom hline
\multirow{2}{*}{$\sigma$} & \multicolumn{4}{c}{APPD (mm)}  \\
                   & mean  & std   & max   & Q95   \\
\noalign{\vskip 2pt} % Adds space near the bottom hline

\hline
\noalign{\vskip 2pt} % Adds space near the bottom hline
1       & \textbf{0.817} & 0.188   & \textbf{1.363}   & \textbf{1.149} \\
2       & 0.842          & 0.243   & 2.289            & 1.195          \\
3       & 0.899          & 0.366   & 2.660            & 1.531          \\
\hline
\multicolumn{5}{l}{Note: UNet-Disp results (lower is better)}
\end{tabular}
\end{table}

\section{Discussion}

The UNet-MLP model exhibited the lowest accuracy, most likely due to two reasons: (1) CNN layers have limited receptive field sizes and (2) displacement predictions on points are made independently without considering possible correlations. 
Compared to UNet-MLP, the UNet-GCN model had relatively better accuracy, mostly because the GCN layers predict the displacement of a point by aggregating information from neighbour points on the template because points in a neighborhood are correlated. But, the UNet-GCN model still had much larger errors in some cases, especially in the L1 and D1 regions, which indicates the limitation of GCN. 
The UNet-Disp model had enhanced accuracy, which is achieved by predicting a displacement field on a regular grid. A major drawback of the UNet-Disp model is that it is not robust to template initialization. As shown by large max and Q95 values in Table \ref{tab:appd-overview} and the examples in Fig. \ref{fig:abnormal-cases}, all of the three models generate irregular shapes in some cases.

Our two models, UNet-DeformSA and TransDeformer significantly (p-value $<$ 1e-6 using paired t-test) outperformed the other models as shown by the much lower mean, max, and Q95 values in Table \ref{tab:appd-overview} and the artifact-free geometries in Fig. \ref{fig:abnormal-cases}. The novel SSA module in UNet-DeformSA is the key to achieve such high performance by considering long-range dependencies among points of the template. By using two novel attention modules (SIA and S2IA) to utilize long-range dependencies among template points and image patches, TransDeformer no longer relies on a UNet for feature extraction and performed better than UNet-DeformSA (p-value $<$ 1e-6 using paired t-test).

All of the current ML models for geometry reconstruction, including our models, are data-driven and therefore have no guarantee in output accuracy for an input image. Thus, the geometry output from a ML model has to be checked (and modified if necessary) by a human operator to ensure high accuracy for clinical use. The shape error estimation model, which is a modified version of TransDeformer, facilitates this quality control process by ranking the model-reconstructed geometries by the estimated errors. To reduce human negligence, those with larger (estimated) errors will be checked first when the operator is vigilant; and those with smaller (estimated) errors will be checked later when the operator possibly becomes fatigue.

Our study used the mid-sagittal lumbar spine MR images for two major reasons. Firstly, as shown in clinical studies \cite{b11, b50}, the mid-sagittal image of a patient provides the most useful information for lumbar spine degeneration assessment. Secondly, the slice thickness of a lumbar MR scan in the sagittal direction is often much larger than 5mm, causing difficulties to create accurate 3D ground-truth annotation. Our model could be directly extended to handle 3D images once the sagittal slice thickness becomes acceptably small with the advancement of imaging technology.

\section{Conclusion}
We proposed two novel attention-based neural networks, UNet-DeformSA and TransDeformer, to automatically reconstruct lumbar spine geometries with mesh correspondence from 2D MR images. The reconstructed geometries are highly accurate and free of artifacts. In addition, we proposed a shape error estimation network based on TransDeformer, which facilitates quality control. Thus, we have provided a complete solution for fast and accurate measurement of the key medical parameters of lumbar spine components from MR images.

\appendices

%\section*{Acknowledgment}
%The preferred spelling of the word ``acknowledgment'' in American English is 
%without an ``e'' after the ``g.'' Use the singular heading even if you have 
%many acknowledgments. Avoid expressions such as ``One of us (S.B.A.) would 
%like to thank $\ldots$ .'' Instead, write ``F. A. Author thanks $\ldots$ .'' In most 
%cases, sponsor and financial support acknowledgments are placed in the 
%unnumbered footnote on the first page, not here.

%\bibliographystyle{unsrt}
\bibliographystyle{IEEEtran}
\bibliography{reference}

\end{document}